\documentclass[letterpaper]{article}
\usepackage{aaai2027}
\usepackage[hyphens]{url}
\usepackage{graphicx}
\usepackage{natbib}
\usepackage{caption}
\usepackage{algorithm}
\usepackage{booktabs}
\usepackage{multirow}
\usepackage{makecell}
\usepackage[table]{xcolor}
\usepackage{graphicx}
\usepackage{xcolor}
\usepackage{subcaption}
\usepackage{amsmath}
\usepackage{amssymb}
\usepackage{url}
\usepackage[most]{tcolorbox}
\usepackage{xcolor}
\usepackage{enumitem}
\usepackage{url}
\usepackage{algpseudocode}
\usepackage{tabularx}
\usepackage{array}
\usepackage{ragged2e}
\usepackage{booktabs}

\newcolumntype{Y}{>{\RaggedRight\arraybackslash}X}
\definecolor{prompttitle}{RGB}{195,195,195}
\definecolor{promptgreen}{RGB}{218,255,218}
\definecolor{promptyellow}{RGB}{255,249,203}
\definecolor{promptred}{RGB}{255,220,220}
\definecolor{promptblue}{RGB}{220,220,255}
\definecolor{promptorange}{RGB}{224,145,65}

\newtcolorbox{freshpromptbox}[1]{
    enhanced,
    breakable,
    width=\columnwidth,
    colback=white,
    colframe=gray!45,
    boxrule=0.5pt,
    arc=0pt,
    outer arc=0pt,
    left=7pt,
    right=7pt,
    top=7pt,
    bottom=7pt,
    title={#1},
    colbacktitle=prompttitle,
    coltitle=black,
    fonttitle=\bfseries\small,
    boxed title style={
        sharp corners,
        boxrule=0pt
    },
    before skip=6pt,
    after skip=6pt
}

\newcommand{\promptsection}[2]{%
    \par\medskip
    \noindent
    \tcbox[
        on line,
        colback=#1,
        colframe=#1,
        boxrule=0pt,
        arc=0pt,
        left=3pt,
        right=3pt,
        top=1.5pt,
        bottom=1.5pt
    ]{\bfseries #2}
    \par\smallskip
}

\usepackage{listings}

\lstdefinestyle{freshcode}{
    basicstyle=\ttfamily\scriptsize,
    columns=fullflexible,
    keepspaces=true,
    breaklines=true,
    breakatwhitespace=false,
    showstringspaces=false,
    frame=none,
    xleftmargin=3pt,
    xrightmargin=3pt,
    aboveskip=3pt,
    belowskip=3pt
}

\algrenewcommand\algorithmicrequire{\textbf{Input:}}
\algrenewcommand\algorithmicensure{\textbf{Output:}}

\newcommand{\impr}[1]{\textbf{\textcolor{green}{#1}}}

\definecolor{bestcolor}{HTML}{E7E4FF}
\definecolor{gracecolor}{HTML}{FCE8CC}
\newcommand{\best}[1]{\cellcolor{bestcolor}\textbf{#1}}
\newcommand{\second}[1]{\cellcolor{bestcolor!45}\underline{#1}}

\definecolor{groupcolor}{HTML}{EEEEEE}
\definecolor{freshcolor}{HTML}{E9F5EC}
\newcommand{\method}{FRESH}
\newcommand{\graph}{\mathcal{G}}
\newcommand{\hist}{\mathcal{H}}
\newcommand{\tools}{\mathcal{T}}

\title{Learning from Failures: Heterogeneous Graph Memory for Small Language Model Tool-Using Agents}
\author{
    Jiaxing Li,\textsuperscript{\rm 1}
    Lei Song,\textsuperscript{\rm 1}
    Rui Dong,\textsuperscript{\rm 1}
    Youyong Kong\textsuperscript{\rm 1}\thanks{Corresponding author.}
}

\affiliations{
    \textsuperscript{\rm 1}School of Computer Science and Engineering, Southeast University, Nanjing, China\\
    \{jiaxing\_li, 230238577, dongrui, kongyouyong\}@seu.edu.cn
}

\begin{document}

\maketitle

\begin{abstract}
Small and medium-sized language models offer cost-effective executors for tool-using agents, making them attractive for local and large-scale deployment. However, in long-horizon and stateful environments, they often make structural errors such as missing required observations, performing premature writes, repeating failed calls, and violating action preconditions. These errors can lead to incorrect state updates, policy violations, and costly or irreversible consequences, making reliable tool execution a critical deployment challenge. Existing fine-tuning approaches require substantial data and computation, while flat memory may retrieve failed actions without preserving their causal context or safety conditions.
In this paper, we propose \textbf{FRESH}, a \textbf{F}ailure-aware \textbf{R}etrieval framework over \textbf{E}xperience-\textbf{S}tructured \textbf{H}eterogeneous graphs, which transforms historical successes and failures into structured external experience for tool-using agents. By explicitly modeling the dependencies among tasks, actions, errors, repairs, and execution conditions, FRESH helps frozen language models reuse reliable strategies, avoid recurring failures, and make safer decisions in stateful tool interactions. Experiments on $\tau$-Bench and AppWorld with multiple open-source models show that FRESH consistently improves task success and tool-use reliability over no-memory agents and representative memory-based baselines.

\end{abstract}

\section{Introduction}

Large language models (LLMs) have demonstrated strong capabilities in reasoning, planning, and interactive decision making. However, many real-world tasks cannot be completed through language generation alone and require access to external tools, such as search engines, databases, calculators, code interpreters, and software APIs \citep{yao2023react,schick2023toolformer,qin2024toolllm}. This motivates LLM-based agents that use the language model as a central controller to interpret user requests, select tools, process observations, and iteratively interact with external environments.

Reliable tool use becomes substantially more difficult in multi-turn and stateful settings. An agent must not only select a syntactically valid tool, but also collect sufficient evidence, maintain
intermediate states, recover from tool errors, follow domain policies, and determine whether a state-changing operation is safe \citep{yao2024taubench,trivedi2024appworld}. These requirements are
particularly challenging for small and medium-sized language models. Although compact models offer lower inference and deployment overhead and are attractive for local or large-scale serving
\citep{liu2024mobilellm}, they are more likely to skip required observations, repeat ineffective calls, lose track of intermediate states, or perform write actions before their preconditions are
satisfied. Such structural errors can result in incorrect database updates, policy violations, or irreversible consequences.

\begin{figure}[t]
\centering
\includegraphics[width=\linewidth]{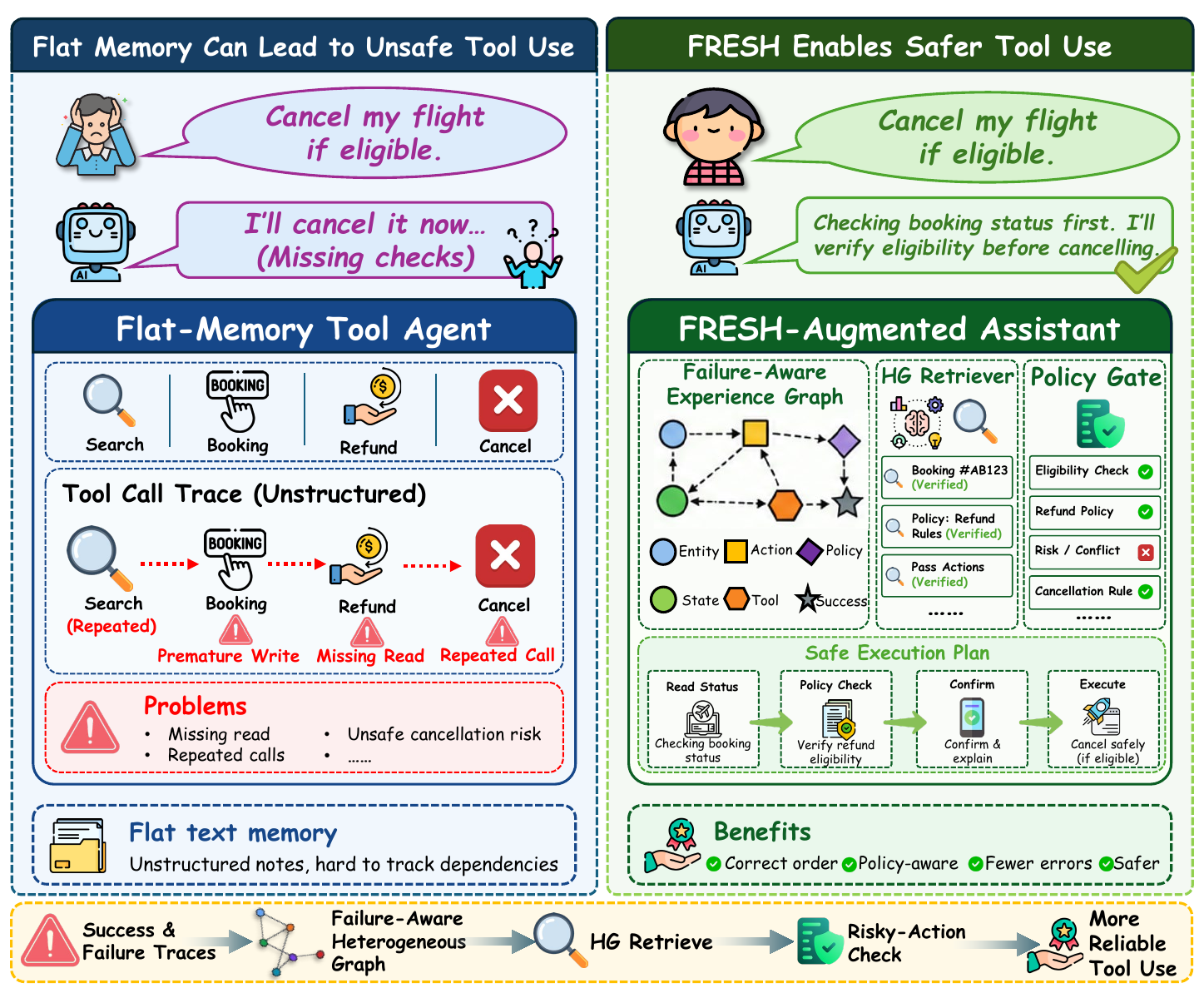}
\vspace{-0.7cm}
\caption{Flat trajectory memory can obscure action dependencies and
failure causes. \method{} structures successful and failed experience
as a heterogeneous graph, retrieves relevant evidence, and checks
risky actions before execution.}
\label{fig:motivation}
\end{figure}

Neither increasing the interaction budget nor repeatedly fine-tuning the executor fully addresses this problem. Additional turns may cause agents to repeat an unproductive strategy without identifying the
missing evidence, while continuously adapting model parameters to changing tools and policies requires substantial training data and computation. Moreover, knowledge encoded in model parameters is
difficult to inspect when new failure patterns are discovered. This motivates an external experience mechanism that can improve execution while keeping the underlying language model frozen.
External memory provides a natural way to reuse knowledge from previous interactions. Existing methods store and retrieve demonstrations, trajectories, reflections, or distilled reasoning
strategies \citep{shinn2023reflexion,park2023generative, ouyang2026reasoningbank}. Recent approaches further introduce graph-structured memory to capture relations among historical experiences or recurring tool transitions \citep{li2026hepm,feng2026expgraph}. Nevertheless, existing memory
mechanisms remain insufficient for policy-sensitive and state-changing tool use \cite{mem02024, ouyang2026reasoningbank}. Flat memories and complete trajectories often leave action dependencies implicit, while success-oriented memories provide limited representations of failure causes, repairs, and write-action preconditions. Consequently, a failed action may be retrieved as a reusable example, or a previously successful write may be copied without the observations and policy conditions that made it valid.

Figure~\ref{fig:motivation} illustrates this distinction. Flat memory \cite{mem02024,li2026hepm} preserves historical content but does not explicitly indicate which observations are required before an action, why a previous action
failed, or how the failure should be repaired. As a result, an agent may repeat a search, skip an eligibility check, or perform a premature cancellation even after retrieving a superficially similar
trajectory. Reliable experience reuse should instead capture what succeeded, what failed, why it failed, how it was repaired, and under which conditions an action is safe.

To address this problem, we propose \method{} (\textbf{F}ailure-aware \textbf{R}etrieval over \textbf{E}xperience-\textbf{S}tructured \textbf{H}eterogeneous graphs), an external experience framework for reliable tool use with frozen small and medium-sized language models. The key idea is to organize historical successes and failures as structured, role-aware experience rather than treating all past actions as equally reusable examples. \method{} represents tasks, tools, actions, observations, errors, repairs, and execution preconditions in a heterogeneous graph, explicitly separating verified strategies from unsafe behaviors. For a new task, it retrieves a relevant candidate subgraph and uses a trained heterogeneous graph retriever to rank the most useful evidence. Instead of replaying complete trajectories, the retrieved experience is compressed into concise guidance on what to perform, avoid, check, or repair, while a lightweight policy and precondition gate validates risky actions against live tool observations and graph-derived constraints before execution. This design provides an inspectable and updateable execution scaffold without modifying the underlying language model. We evaluate \method{} on $\tau$-bench and AppWorld with multiple open-source small and medium-sized models, and the results show consistent improvements in task completion and tool-use reliability over no-memory, prompting, retrieval, trajectory, and experience-memory baselines, particularly in long-horizon and policy-sensitive settings.
Our main contributions are summarized as follows:
\begin{itemize}
    \item We formulate reliable experience reuse for tool-using agents as a failure-aware memory organization problem, explicitly distinguishing successful strategies, failed actions, repair
    patterns, and write-action preconditions.

    \item We propose an experience-structured heterogeneous graph that captures the dependencies among tasks, tools, actions, observations, errors, repairs, and execution conditions.

    \item We develop a graph-aware retrieval and execution framework that produces compact operational guidance and constrains risky tool actions through a lightweight policy and precondition gate.

    \item We evaluate \method{} on $\tau$-bench and AppWorld across multiple open-source small and medium-sized models, demonstrating consistent improvements over representative memory-based
    baselines.
\end{itemize}
\section{Related Work}

\paragraph{Tool-use agents and benchmarks.}
Language-model agents extend text generation with planning, tool selection, API invocation, and interaction with external environments. Early work studies tool-augmented reasoning and API learning through language-model prompting or supervision \citep{karpas2022mrkl,schick2023toolformer,patil2024gorilla,qin2024toolllm}, while ReAct interleaves reasoning and actions during task execution \citep{yao2023react}. Evaluation has consequently progressed from isolated function calls to stateful, long-horizon interaction. BFCL evaluates function selection, argument generation, and multi-turn function calling over realistic schemas \citep{patil2025bfcl}. tau-bench evaluates conversations among an agent, a simulated user, and stateful tools under domain policies \citep{yao2024taubench}. AppWorld further requires agents to operate applications through APIs and evaluates the resulting environment state with hidden tests \citep{trivedi2024appworld}. These benchmarks reveal that reliable tool use requires more than syntactically valid calls: agents must preserve state, satisfy action dependencies, recover from errors, and avoid unsafe or premature operations.

\paragraph{Experience memory and agent adaptation.}
Existing approaches improve tool-use agents by storing experience or adapting the agent pipeline \cite{packer2023memgpt}. Reflexion and Voyager reuse textual reflections or skills from previous interactions \citep{shinn2023reflexion,wang2023voyager}. ReasoningBank distills successful and failed trajectories into reusable reasoning memories \citep{ouyang2026reasoningbank}, while Mem0 stores concise long-term memories for subsequent semantic retrieval \citep{mem02024}. H-EPM organizes successful episodes into episodic and procedural memories with tool-transition structure \citep{li2026hepm}. PACE instead improves a frozen model through two-timescale prompt and control-logic evolution, accepting structural changes through held-out validation \citep{ling2026pace}. Despite their effectiveness, existing memory methods either provide limited dependency structure or insufficiently model failures, repairs, and write preconditions, leaving final execution decisions largely to the language model. In contrast, \method{} organizes these signals in a typed heterogeneous graph, performs relation-aware retrieval, and converts retrieved evidence into compact guidance and explicit checks for risky actions.

\begin{figure*}[t]
\centering
\includegraphics[width=\textwidth]{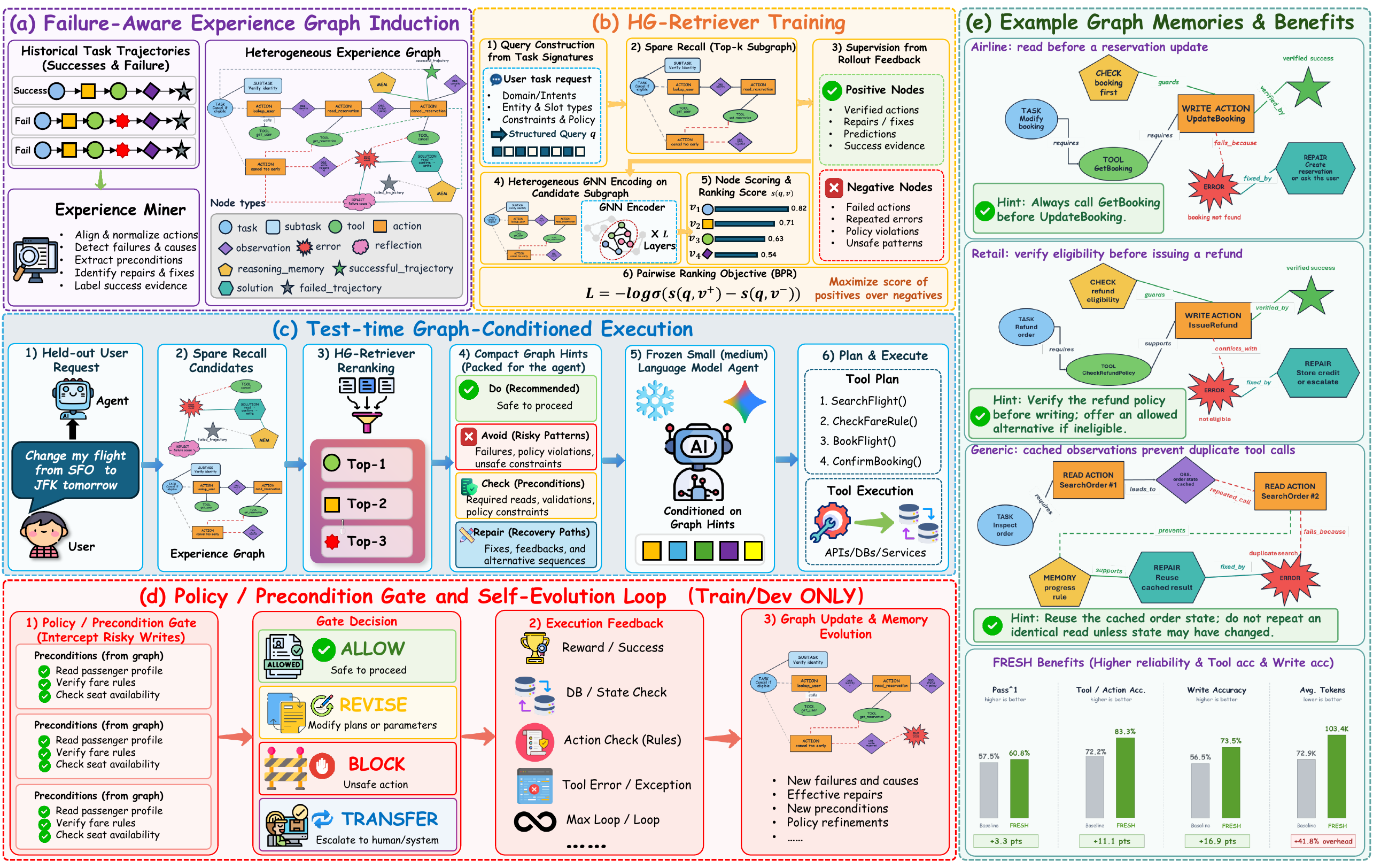}
\vspace{-0.7cm}
\caption{Overview of \method{}. During training and development, successful and failed tool-use trajectories are transformed into a heterogeneous experience graph containing actions, tools, observations, errors, preconditions, and repairs. A feedback-supervised heterogeneous graph retriever is trained to rank useful experience within sparsely recalled candidate subgraphs. At test time, the frozen graph and retriever produce compact \emph{do}, \emph{avoid}, \emph{check}, and \emph{repair} guidance for a frozen language-model executor. A policy and precondition gate validates risky actions before execution. The feedback and graph-evolution loop is active only during training and development, while the memory remains frozen during evaluation.}
\label{fig:overview}
\end{figure*}

\section{Method}
\subsection{Problem Setup}

We study interactive tool-use tasks in which an agent completes a user request $x$ by interacting with external APIs, applications, or databases. At step $t$, the agent observes the interaction history $h_t$ and either returns a response, terminates, or invokes a tool $a_t=(u_t,z_t)$, where $u_t\in\tools$ is a tool and $z_t$ contains its arguments. The environment returns an observation $o_t$ and may update its internal state. Tools may retrieve information, modify persistent state, or complete part of a multi-application workflow. The resulting trajectory $\tau=(x,a_1,o_1,\ldots,a_T,o_T)$ receives benchmark-specific feedback $F$, such as task success, evaluator tests, state checks, tool errors, or termination status.

Our goal is to improve a fixed executor $\pi_\theta$ without updating its language-model parameters. Given successful and failed training trajectories $\hist_{\mathrm{train}}$, \method{} constructs a heterogeneous experience graph $\graph_{\mathrm{train}}$ and trains an auxiliary graph retriever. Both the graph and retriever are frozen during evaluation to prevent information leakage from held-out test tasks.
At step $t$, the executor is conditioned on the current history and retrieved graph evidence:
\[
    \hat{a}_t \sim \pi_\theta
    \bigl(a \mid h_t, R_{\graph}(h_t)\bigr),
\]
where $R_{\graph}(h_t)$ contains compact success patterns, failure warnings, preconditions, and repairs. For actions requiring additional validation, an execution gate maps the proposed action to \textsc{allow}, \textsc{revise}, \textsc{block}, or \textsc{transfer}. We evaluate using $\mathrm{Pass}^k$ and action metrics on $\tau$-bench, task pass rate and evaluator-test completion on AppWorld, together with execution steps and inference cost.
\subsection{Overall Framework}
Figure~\ref{fig:overview} presents the overall architecture of \method{}. The framework separates offline experience learning from test-time tool execution. During training and development, \method{} collects both successful and failed trajectories and converts them into a heterogeneous experience graph. The graph explicitly represents tasks, tools, actions, observations, errors, preconditions, successful outcomes, and repair strategies, allowing the system to preserve both reusable behavior and failure causes.
To retrieve useful experience efficiently, \method{} first applies sparse recall to identify a query-relevant candidate subgraph. A trained heterogeneous GNN then propagates information over typed nodes and relations and reranks the candidates using supervision derived from trajectory feedback. The selected evidence is compressed into four types of operational guidance: recommended actions, patterns to avoid, conditions to check, and possible repair strategies.
At test time, the language-model executor remains frozen and generates actions conditioned on the current interaction history and retrieved graph guidance. Before an action with potential side effects is executed, a policy and precondition gate checks it against graph-derived constraints and current tool observations. The gate can allow the action, revise its plan or arguments, block an unsafe operation, or transfer the task when automatic execution is inappropriate. Execution feedback is used to evolve the graph only during training and development; both the experience graph and retriever are frozen on held-out test tasks.

\subsection{Failure-Aware Experience Graph Induction}

For each task request $x_i$, the agent produces a tool-use trajectory
\begin{equation}
    \tau_i =
    \left(
        x_i,
        a_{i,1}, o_{i,1},
        \ldots,
        a_{i,T_i}, o_{i,T_i},
        F_i
    \right),
    \label{eq:trajectory}
\end{equation}
where $a_{i,t}$ and $o_{i,t}$ denote the action and observation at step $t$, and $F_i$ contains environment feedback such as task reward, state and action checks, tool errors, repeated calls, and termination status. The experience miner normalizes tools, arguments, entities, and observations, and performs action-level credit assignment to distinguish verified actions from failure patterns. For each failed action, it further extracts the failure cause, missing precondition, and corresponding repair, rather than treating the entire failed trajectory as incorrect.

The extracted experience is organized as a typed directed heterogeneous graph
\begin{equation}
    \mathcal{G}
    =
    \left(
        \mathcal{V},
        \mathcal{E},
        \tau_{\mathcal{V}},
        \tau_{\mathcal{E}},
        \mathbf{X}
    \right),
    \label{eq:experience_graph}
\end{equation}
where $\mathcal{V}$ and $\mathcal{E}$ are nodes and edges, $\tau_{\mathcal{V}}$ and $\tau_{\mathcal{E}}$ denote their types, and $\mathbf{X}$ contains textual and statistical attributes. Nodes represent tasks, tools, actions, observations, errors, preconditions, repairs, reasoning memories, and successful or failed trajectories, while relations such as \texttt{calls}, \texttt{requires}, \texttt{guards}, \texttt{verified\_by}, \texttt{fails\_because}, and \texttt{fixed\_by} capture their dependencies. By separating unsafe actions from the warnings, preconditions, and repairs derived from them, the graph preserves useful failure knowledge without encouraging the agent to repeat failed behaviors. For example, if an agent correctly reads a booking but attempts to cancel it before verifying its eligibility, the read action is retained as verified evidence, whereas the premature cancellation is marked as a failure. The miner further creates an error node describing the missing eligibility check, a precondition node that guards the cancellation, and a repair node representing the corrected read--check--confirm--write sequence. We provide a detailed example in Appendix~A.

\subsection{Heterogeneous Graph Retriever Training}
The HG-Retriever is trained offline to rank useful graph memories for a given task. As shown in Figure~\ref{fig:overview}, the training process contains query construction, sparse candidate recall, feedback-derived supervision, heterogeneous graph encoding, and pairwise ranking.

\paragraph{Query construction and sparse recall.}
For each task, we construct a structured query representation from its request and interaction context:
\begin{equation}
    \mathbf{q}_i = f_q(x_i,h_i),
    \label{eq:query_encoding}
\end{equation}
where the input includes domain and intent cues, entities and slot types, available tools, policy constraints, and recent error
information. A sparse retriever first selects the top-$K_0$ nodes according to textual relevance, node type, and historical utility. Their local neighbors are then included to form a compact candidate subgraph
\begin{equation}
    \mathcal{G}_i^{c} = \mathcal{G}
    \left[
        \operatorname{TopK}_{K_0}(q_i) \cup \mathcal{N}^{H}
        \left(
            \operatorname{TopK}_{K_0}(q_i)
        \right)
    \right],
    \label{eq:candidate_graph}
\end{equation}
where $\mathcal{N}^{H}(\cdot)$ denotes typed neighbors within $H$ hops. Sparse recall limits GNN computation to a small task-relevant region instead of encoding the complete memory graph for every query.

\paragraph{Feedback-derived supervision.}
Positive nodes are automatically obtained from verified actions, successful reasoning memories, effective repairs, and preconditions that support successful writes. Negative nodes include failed actions, repeated calls, policy violations, irrelevant memories, and unsafe execution patterns. Importantly, a reflection such as ``check eligibility before refunding'' is treated as useful guidance, whereas the raw premature refund action is treated as a negative example. Thus, supervision is derived from environment feedback without requiring manual relevance annotation. 

\paragraph{Heterogeneous graph encoding.}
The initial representation of candidate node $v$ combines its textual
content, node type, historical statistics, and query-matching
features:
\begin{equation}
    \mathbf{h}_{v}^{(0)}
    =
    W_{\mathrm{in}}
    \left[
        \mathbf{e}_{v}^{\mathrm{text}}
        \,\Vert\,
        \mathbf{e}_{v}^{\mathrm{type}}
        \,\Vert\,
        \mathbf{e}_{v}^{\mathrm{stat}}
        \,\Vert\,
        \mathbf{e}_{i,v}^{\mathrm{match}}
    \right],
    \label{eq:node_features}
\end{equation}
where $\mathbf{e}_{v}^{\mathrm{text}}$ denotes the textual embedding,
$\mathbf{e}_{v}^{\mathrm{type}}$ denotes the node-type embedding,
$\mathbf{e}_{v}^{\mathrm{stat}}$ contains historical success and
failure statistics, and $\mathbf{e}_{i,v}^{\mathrm{match}}$ measures
the compatibility between query $q_i$ and node $v$.
Here, $\Vert$ denotes concatenation and $W_{\mathrm{in}}$ projects the
combined features into a shared hidden space.

We apply relation-aware message passing over the candidate subgraph:
\begin{equation}
    \mathbf{h}_{v}^{(\ell+1)}
    =
    \sigma
    \left(
        W_{0}^{(\ell)}\mathbf{h}_{v}^{(\ell)}
        +
        \sum_{r\in\mathcal{R}}
        \frac{1}{|\mathcal{N}_{r}(v)|}
        \sum_{u\in\mathcal{N}_{r}(v)}
        W_{r}^{(\ell)}\mathbf{h}_{u}^{(\ell)}
    \right),
    \label{eq:hetero_message_passing}
\end{equation}
where $\mathcal{N}_{r}(v)$ is the set of neighbors connected to $v$ through relation $r$, and $W_r^{(\ell)}$ is a relation-specific transformation. Consequently, an action node can incorporate
information from its associated tool, observation, precondition, error, and repair nodes. After $L$ layers, the query-conditioned relevance score is computed as
\begin{equation}
    s_{\phi}(q_i,v)
    =
    \operatorname{MLP}
    \left(
        \mathbf{h}_{v}^{(L)}
        \,\Vert\,
        \mathbf{q}_{i}
        \,\Vert\,
        \mathbf{h}_{v}^{(L)}
        \odot
        \mathbf{q}_{i}
    \right),
    \label{eq:node_score}
\end{equation}
where $\phi$ denotes the retriever parameters and $\odot$ denotes element-wise interaction.

\paragraph{Training objective.}
For each query, we sample a positive node $v^{+}$ and a negative node
$v^{-}$ from its candidate subgraph. The retriever is optimized using
a pairwise ranking objective:
\begin{equation}
    \mathcal{L}_{\mathrm{rank}}
    =
    -
    \sum_{i}
    \log
    \sigma
    \left(
        s_{\phi}(q_i,v^{+})
        -
        s_{\phi}(q_i,v^{-})
    \right).
    \label{eq:ranking_loss}
\end{equation}
This objective encourages verified actions, useful preconditions, and
effective repairs to receive higher scores than failed actions,
repeated calls, policy violations, and irrelevant memories. The final
training objective is
\begin{equation}
    \mathcal{L}_{\mathrm{HG}}
    =
    \mathcal{L}_{\mathrm{rank}}
    +
    \lambda_{\mathrm{reg}}
    \|\phi\|_{2}^{2},
    \label{eq:total_retriever_loss}
\end{equation}
where $\phi$ denotes the parameters of the HG-Retriever. Only the
retriever is optimized, while the executor language model remains
frozen.

\subsection{Test-time Graph-Conditioned Execution}

At test time, the experience graph and trained HG-Retriever are
frozen. Given the current user request and interaction history,
\method{} performs the same sparse recall and GNN reranking process
and selects the top-ranked graph nodes. These nodes are organized into
four operational categories:
\begin{itemize}
    \item \textbf{Do}: verified actions or tool-use patterns;
    \item \textbf{Avoid}: failed actions, repeated calls, and unsafe
    patterns;
    \item \textbf{Check}: required observations, validations, and
    policy conditions;
    \item \textbf{Repair}: corrections, fallback tools, and
    alternative execution paths.
\end{itemize}

The selected evidence is compressed under a fixed context budget:
\begin{equation}
    M_t =
    \operatorname{Serialize}
    \left(
        \mathcal{R}_{t}^{\mathrm{Do}},
        \mathcal{R}_{t}^{\mathrm{Avoid}},
        \mathcal{R}_{t}^{\mathrm{Check}},
        \mathcal{R}_{t}^{\mathrm{Repair}}
    \right),
    \label{eq:memory_serialization}
\end{equation}
where $\operatorname{Serialize}(\cdot)$ converts the retrieved graph
nodes into a compact textual prompt organized under the
\textsc{Do}, \textsc{Avoid}, \textsc{Check}, and \textsc{Repair}
categories. $M_t$ is a compact memory sheet rather than a complete historical trajectory. This representation removes task-specific identifiers and irrelevant observations while preserving actionable dependencies. The frozen executor generates a candidate action according to
\begin{equation}
    \widetilde{a}_{t}
    \sim
    \pi_{\theta}
    \left(
        a
        \mid
        h_t,
        M_t,
        \mathcal{T}
    \right),
    \qquad
    \theta\ \text{is frozen},
    \label{eq:test_execution}
\end{equation}
where $h_t$ is the current history and $\mathcal{T}$ contains the available tool schemas. Low-risk read actions can be executed directly, whereas state-changing, terminating, or policy-sensitive actions are passed to the policy and precondition gate.

\subsection{Policy / Precondition Gate and Self-Evolution Loop} 
The policy and precondition gate prevents retrieved experience from being copied without considering the current environment state. For a candidate action $\widetilde{a}_t$, the gate considers its required preconditions $\mathcal{P}_t$, live observations $\mathcal{O}_t$, and domain policies $\mathcal{B}_t$:
\begin{equation}
    d_t
    =
    g
    \left(
        \widetilde{a}_t,
        \mathcal{P}_t,
        \mathcal{O}_t,
        \mathcal{B}_t
    \right)
    \in
    \{
        \mathrm{Allow},
        \mathrm{Revise},
        \mathrm{Block},
        \mathrm{Transfer}
    \}.
    \label{eq:gate}
\end{equation}

The action is allowed when all required observations and policy conditions are verified. It is revised when the action is potentially valid but requires a missing read, corrected argument, or alternative tool sequence. The gate blocks operations that violate hard constraints and transfers tasks that require human intervention or
cannot be safely resolved from the available evidence. During training and development, execution feedback is returned to the experience miner. Successful actions increase the utility of their supporting memories, whereas tool errors, repeated calls, failed state checks, and blocked writes reduce their utility:
\begin{equation}
    u_v^{(t+1)}
    =
    (1-\beta)u_v^{(t)}
    +
    \beta\,
    \Delta(F_t,v),
    \label{eq:utility_update}
\end{equation}
where $\beta\in[0,1]$ controls the update rate of the historical utility,
and $\Delta(F_t,v)$ denotes the feedback assigned to memory node $v$
according to the execution result $F_t$. Newly observed failure causes, repairs, and preconditions are inserted into the graph, and the updated graph can be used to generate additional retriever training pairs, which is enabled only during training and development. Once graph construction and retriever training are complete, the final graph and model checkpoint are frozen for evaluation.
\section{Experiments}

\begin{table*}[ht]
\centering
\caption{Experiment results on $\tau$-bench. The shaded bold cells indicate the best results, and underlined cells indicate the second-best results. Avg. denotes the arithmetic mean over all Airline and Retail metrics. FRESH Impr. denotes the relative improvement of FRESH over the base no-memory agent.}
\label{tab:taubench_results}
\vspace{-0.3cm}
\resizebox{\textwidth}{!}{
\begin{tabular}{clccccccccc}
\toprule
\multirow{2}{*}{\textbf{LLM}}&\multirow{2}{*}{\textbf{Method}} 
& \multicolumn{4}{c}{\textbf{Airline}} 
& \multicolumn{4}{c}{\textbf{Retail}}
& \multirow{2}{*}{\textbf{Avg.}} \\
\cmidrule(lr){3-6} 
\cmidrule(lr){7-10}
&& pass$^1$ & pass$^2$ & pass$^3$ & pass$^4$
& pass$^1$ & pass$^2$ & pass$^3$ & pass$^4$
& \\
\midrule

\midrule
\multirow{9}{*}{\rotatebox[origin=c]{90}{Gemma4-26B-A4B}}&
Base
& 0.488 & \second{0.400} & 0.363 & \second{0.350}
& 0.669 & 0.546 & 0.488 & 0.450
& \second{0.469} \\

&Train-Text Prompt 
& \second{0.525} & 0.383 & 0.300 & 0.250
& \second{0.675} & 0.567 & 0.494 & 0.450
& 0.456 \\

&Vector-RAG Memory 
& 0.450 & 0.325 & 0.250 & 0.200
& 0.669 & \second{0.575} & \second{0.525} & \second{0.500}
& 0.437 \\

&ExpeL 
& 0.475 & 0.342 & 0.288 & 0.250
& 0.663 & \second{0.575} & 0.513 & 0.475
& 0.448 \\

&ReasoningBank 
& 0.350 & 0.258 & 0.225 & 0.200
& 0.644 & 0.546 & 0.500 & 0.475
& 0.400 \\

&Mem-0 
& 0.438 & 0.375 & \second{0.363} & \second{0.350}
& 0.656 & 0.562 & 0.512 & 0.475
& 0.466 \\

&H-EPM
& 0.463 & 0.342 & 0.288 & 0.250
& 0.619 & 0.517 & 0.469 & 0.450
& 0.425 \\

&\textbf{FRESH (Ours)} 
& \best{0.538} & \best{0.450} & \best{0.413} & \best{0.400}
& \best{0.700} & \best{0.588} & \best{0.544} & \best{0.525}
& \best{0.520} \\

&Impr. 
& \impr{+10.3\%} & \impr{+12.5\%} & \impr{+13.8\%} & \impr{+14.3\%}
& \impr{+4.7\%} & \impr{+7.6\%} & \impr{+11.5\%} & \impr{+16.7\%}
& \impr{+10.8\%} \\

\midrule
\midrule
\multirow{9}{*}{\rotatebox[origin=c]{90}{Llama3.2-1B-Instruct}}
&Base
& 0.288 & 0.233 & 0.188 & 0.150
& \second{0.100} & \second{0.050} & 0.025 & 0.025
& 0.132 \\

&Train-Text Prompt 
& 0.300 & 0.275 & \second{0.263} & \best{0.250}
& 0.038 & 0.029 & 0.025 & 0.025
& 0.151 \\

&Vector-RAG Memory 
& 0.288 & 0.233 & 0.188 & 0.150
& 0.050 & \second{0.050} & \best{0.050} & \best{0.050}
& 0.132 \\

&ExpeL 
& \second{0.313} & \second{0.283} & \second{0.263} & \best{0.250}
& 0.050 & \second{0.050} & \best{0.050} & \best{0.050}
& \second{0.164} \\

&ReasoningBank 
& 0.300 & 0.275 & \second{0.263} & \best{0.250}
& 0.038 & 0.025 & 0.013 & 0.000
& 0.146 \\

&Mem-0 
& 0.300 & 0.258 & 0.225 & 0.200
& 0.038 & 0.025 & 0.013 & 0.000
& 0.132 \\

&H-EPM
& 0.275 & 0.208 & 0.150 & 0.100
& 0.044 & 0.037 & 0.031 & 0.025
& 0.109 \\

&\textbf{FRESH (Ours)} 
& \best{0.325} & \best{0.300} & \best{0.275} & \best{0.250}
& \best{0.150} & \best{0.071} & \best{0.050} & \best{0.050}
& \best{0.184} \\

&Impr. 
& \impr{+13.0\%} & \impr{+28.6\%} & \impr{+46.7\%} & \impr{+66.7\%}
& \impr{+50.0\%} & \impr{+41.7\%} & \impr{+100.0\%} & \impr{+100.0\%}
& \impr{+38.9\%} \\

\midrule
\midrule
\multirow{9}{*}{\rotatebox[origin=c]{90}{Qwen3-8B}}
&Base
& 0.275 & 0.200 & 0.163 & 0.150
& 0.350 & 0.208 & 0.144 & 0.100
& 0.199 \\

&Train-Text Prompt 
& 0.313 & 0.208 & 0.150 & 0.100
& \best{0.431} & \second{0.317} & 0.250 & 0.200
& 0.246 \\

&Vector-RAG Memory 
& 0.275 & 0.183 & 0.125 & 0.100
& 0.313 & 0.179 & 0.100 & 0.050
& 0.166 \\

&ExpeL 
& \second{0.375} & \second{0.250} & 0.175 & 0.150
& 0.388 & 0.300 & \second{0.256} & \second{0.225}
& \second{0.265} \\

&ReasoningBank 
& 0.300 & 0.233 & \second{0.213} & \second{0.200}
& 0.363 & 0.263 & 0.200 & 0.150
& 0.240 \\

&Mem-0 
& 0.263 & 0.142 & 0.113 & 0.100
& 0.394 & 0.246 & 0.200 & 0.175
& 0.204 \\

&H-EPM
& 0.288 & 0.117 & 0.113 & 0.100
& 0.400 & 0.296 & 0.244 & 0.200
& 0.220 \\

&\textbf{FRESH (Ours)} 
& \best{0.400} & \best{0.267} & \best{0.250} & \best{0.250}
& \second{0.413} & \best{0.325} & \best{0.281} & \best{0.250}
& \best{0.305} \\

&Impr. 
& \impr{+45.5\%} & \impr{+33.4\%} & \impr{+53.4\%} & \impr{+66.7\%}
& \impr{+18.0\%} & \impr{+56.3\%} & \impr{+95.1\%} & \impr{+150.0\%}
& \impr{+53.2\%} \\
\bottomrule
\end{tabular}
}
\end{table*}
\subsection{Experimental Setup}

\subsubsection{Benchmarks.}
We evaluate \method{} on two representative tool-use benchmarks. \textbf{$\tau$-bench} \cite{yao2024taubench} evaluates multi-turn interactions with stateful tools and domain policies in the Airline and Retail domains. We report pass$^k$ for $k\in{1,2,3,4}$ to measure task success across repeated trials. \textbf{AppWorld} \cite{trivedi2024appworld} evaluates long-horizon API interactions across multiple software applications. We report the Pass Rate (PR) on the normal and challenging test splits, together with the average number of execution steps.

\subsubsection{Baselines.}
We compare \method{} with several categories of memory methods, including the no-memory agent, prompt-based memory (\textbf{Train-Text}) \citep{brown2020language}, retrieval-based memory (\textbf{Vector-RAG}) \citep{lewis2020retrieval}, trajectory-based memory (\textbf{Expel}) \citep{zhao2024expel}, and experience-learning memory methods, including \textbf{ReasoningBank} \citep{ouyang2026reasoningbank}, \textbf{Mem0} \citep{mem02024}, and \textbf{H-EPM} \citep{li2026hepm}. All methods use the same task splits, backbone models, execution budgets, and historical trajectories for fair comparison.

\subsubsection{SLM Backbones.}
We conduct experiments with three open-source language models of different capacities: \textbf{Llama-3.2-1B-Instruct}, \textbf{Qwen3-8B} \citep{yang2025qwen3}, and \textbf{Gemma-4-26B-A4B} \citep{gemmateam2026gemma4}. These models represent lightweight, intermediate, and relatively capable tool-use executors, respectively. Their parameters remain frozen throughout the experiments, and \method{} improves their tool-use behavior solely through external structured experience and retrieval. Detailed experimental settings and implementation details are
provided in Appendix~B.

\begin{table}[t]
    \centering
    \caption{Performance comparison on AppWorld. 
    Higher PR is better, while fewer execution steps are preferred.
    The best results are in \best{bold}, and the second-best results
    are \second{underline}.}
    \label{tab:appworld_results}
    \vspace{-0.3cm}
    \resizebox{\columnwidth}{!}{
    \begin{tabular}{lcccc}
        \toprule
        \multirow{2}{*}{Method}
        & \multicolumn{2}{c}{Test-N}
        & \multicolumn{2}{c}{Test-C} \\
        \cmidrule(lr){2-3}
        \cmidrule(lr){4-5}
        & PR $\uparrow$
        & \makecell{\#Steps $\downarrow$}
        & PR $\uparrow$
        & \makecell{\#Steps $\downarrow$} \\
        \midrule

        Gemma4-26B
        & 0.131
        & 10.8
        & 0.043
        & 11.1 \\

        Train-Text Prompt
        & \second{0.250}
        & 10.0
        & 0.082
        & 10.5 \\

        Vector-RAG Memory
        & 0.202
        & \second{9.6}
        & 0.106
        & \second{9.9} \\

        ExpeL
        & 0.220
        & 10.1
        & 0.077
        & 10.8 \\

        ReasoningBank
        & 0.244
        & \best{9.4}
        & 0.110
        & \best{9.7} \\

        Mem0
        & 0.173
        & 10.2
        & 0.079
        & 10.6 \\

        H-EPM
        & \second{0.250}
        & 9.9
        & \second{0.115}
        & 10.4 \\

        \midrule
        \best{FRESH}
        & \best{0.268}
        & \best{9.4}
        & \best{0.146}
        & 10.0 \\

        \bottomrule
    \end{tabular}
    }
\end{table}

\subsection{Performance on $\tau$-bench and AppWorld}
Table~\ref{tab:taubench_results} shows that \method{} achieves the highest overall performance across all three language-model executors. It consistently improves the relatively capable Gemma-4-26B-A4B, provides particularly large gains for the weaker Llama-3.2-1B, and obtains the best average performance with Qwen3-8B despite a few baselines leading on individual metrics. Compared with prompt-based, vector-based, trajectory-based, and experience-learning memories, \method{} is more stable across models and domains because it explicitly distinguishes successful strategies, failure patterns, repairs, and execution preconditions. Table~\ref{tab:appworld_results} further shows that \method{} achieves the highest task-completion rate on both AppWorld splits, while using the fewest steps on Test-N and maintaining competitive execution efficiency on Test-C. These results demonstrate that failure-aware graph memory-based FRESH improves both policy-sensitive conversational tool use and complex long-horizon API interactions.

\subsection{Ablation Study}
\label{sec:ablation}
\begin{figure}[t]
    \centering
    \begin{subfigure}[t]{0.225\textwidth}
        \centering
        \includegraphics[width=\linewidth]{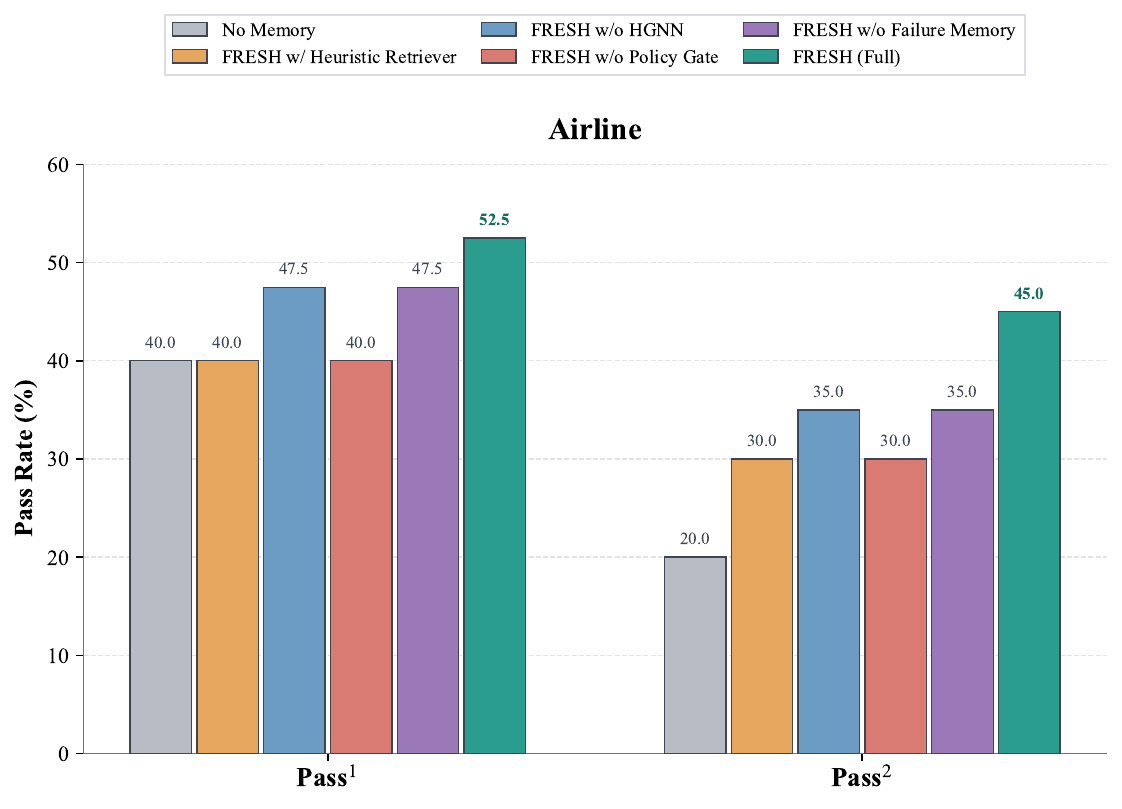}
        \caption{$\tau$-bench Airline}
        \label{fig:airline}
    \end{subfigure}
    \hfill
    \begin{subfigure}[t]{0.225\textwidth}
        \centering
        \includegraphics[width=\linewidth]{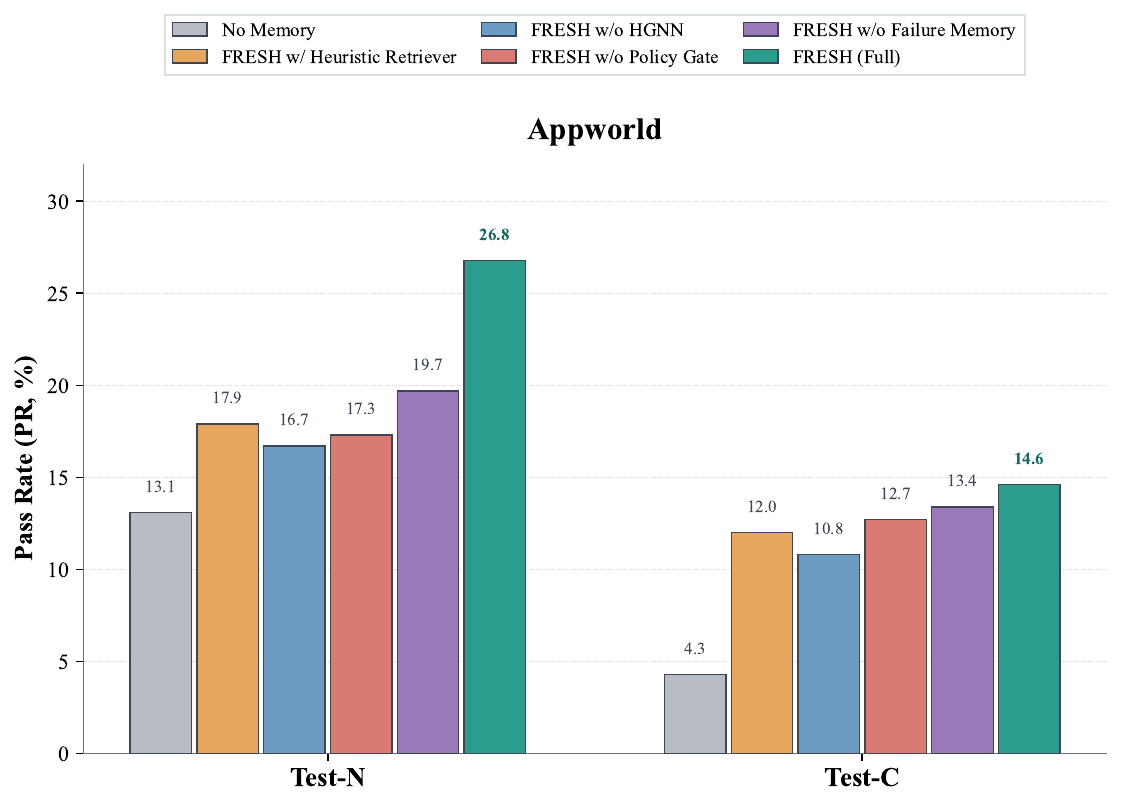}
        \caption{AppWorld}
        \label{fig:appworld}
    \end{subfigure}
    \vspace{-0.3cm}
    \caption{
    Ablation study of FRESH.
    }
    \label{fig:fresh_ablation}
\end{figure}

We conduct ablation studies on $\tau$-bench Airline and AppWorld using five variants: \textbf{No Memory}; \textbf{\method{} w/ Heuristic Retriever}, which replaces the learned retriever with semantic similarity and utility heuristics; \textbf{\method{} w/o HGNN}, which removes heterogeneous graph reranking; \textbf{\method{} w/o Policy Gate}, which disables risky-action validation; and \textbf{\method{} w/o Failure Memory}, which retains only successful experience. Figure~\ref{fig:fresh_ablation} shows that the full \method{} consistently performs best. The policy gate contributes most on Airline, where execution is strongly constrained by action preconditions, while the learned HGNN retriever is particularly important on AppWorld for distinguishing useful strategies from failed or irrelevant experience. Removing failure memory also causes consistent degradation, confirming that failure causes and repair patterns provide complementary guidance beyond successful trajectories. 

\begin{figure}[t]
    \centering
    \includegraphics[width=0.9\linewidth]{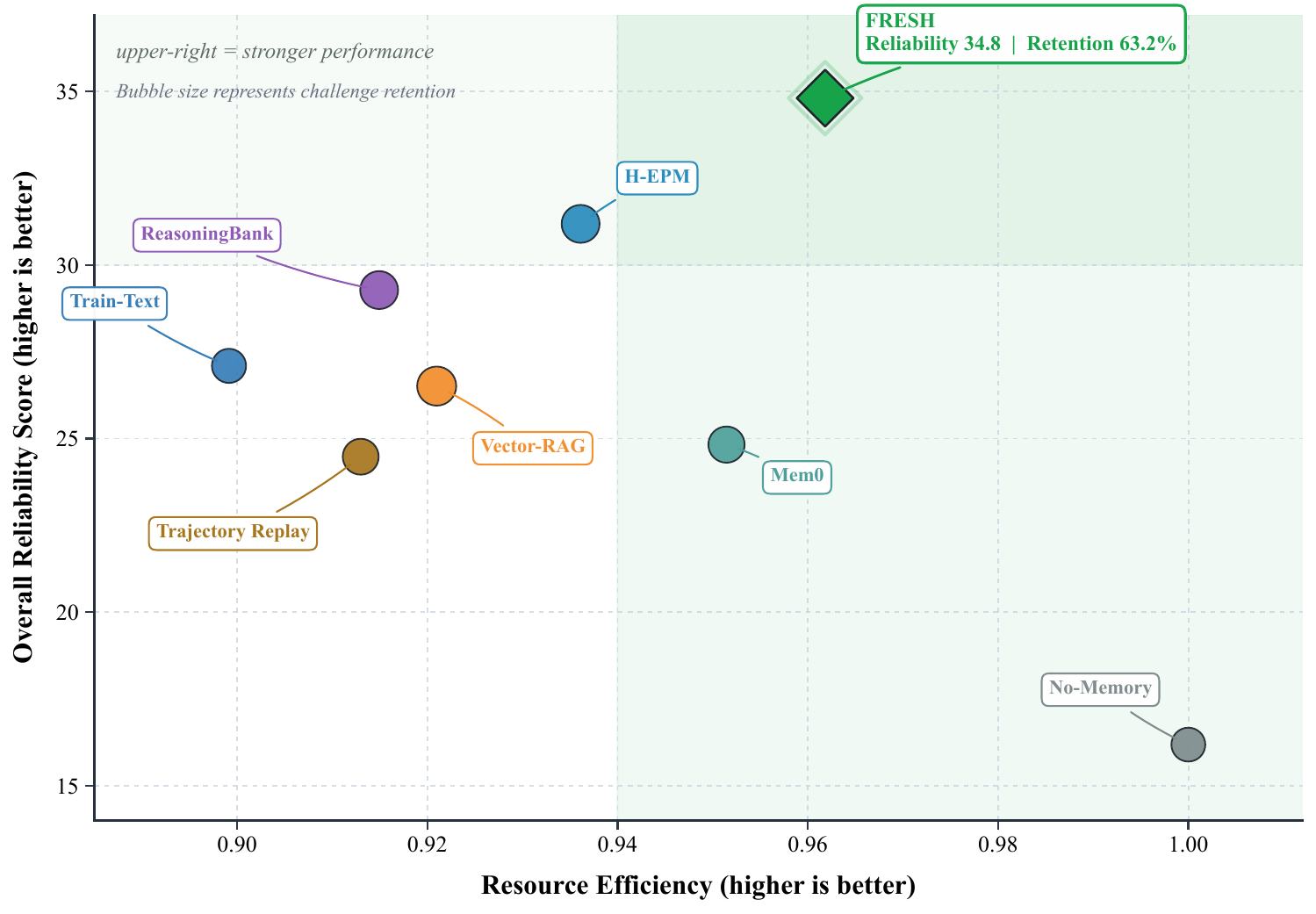}
    \vspace{-0.4cm}
    \caption{Performance--efficiency comparison across memory methods, where the upper-right region indicates a better trade-off between resource efficiency and overall reliability, and bubble size denotes challenge retention.}
    \label{fig:performance--efficiency}
\end{figure}

\begin{figure}[t]
    \centering
    \begin{subfigure}[t]{0.23\textwidth}
        \centering
        \includegraphics[width=\linewidth]{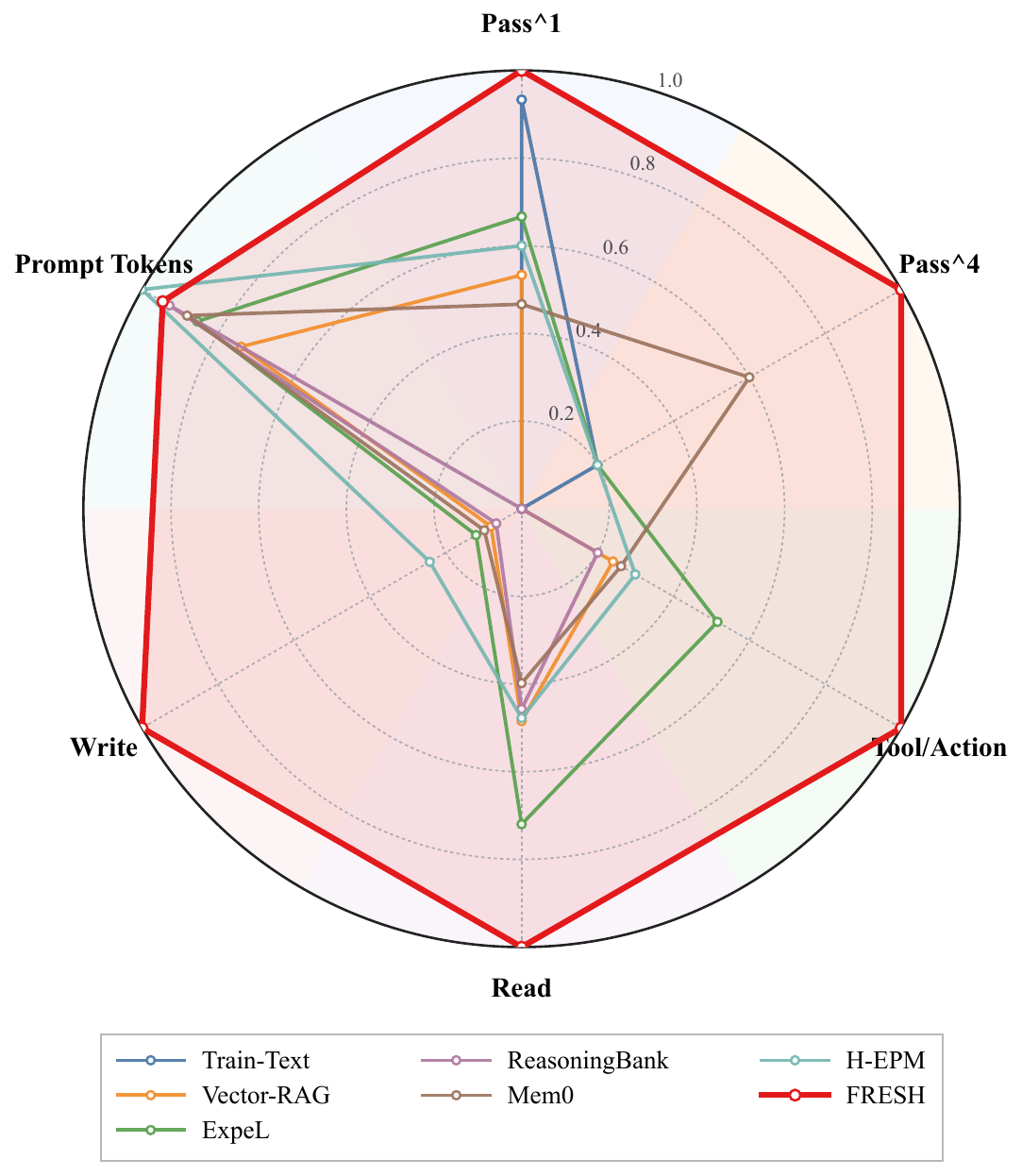}
        \caption{$\tau$-bench Airline}
        \label{fig:radar-tau-airline}
    \end{subfigure}
    \hfill
    \begin{subfigure}[t]{0.23\textwidth}
        \centering
        \includegraphics[width=\linewidth]{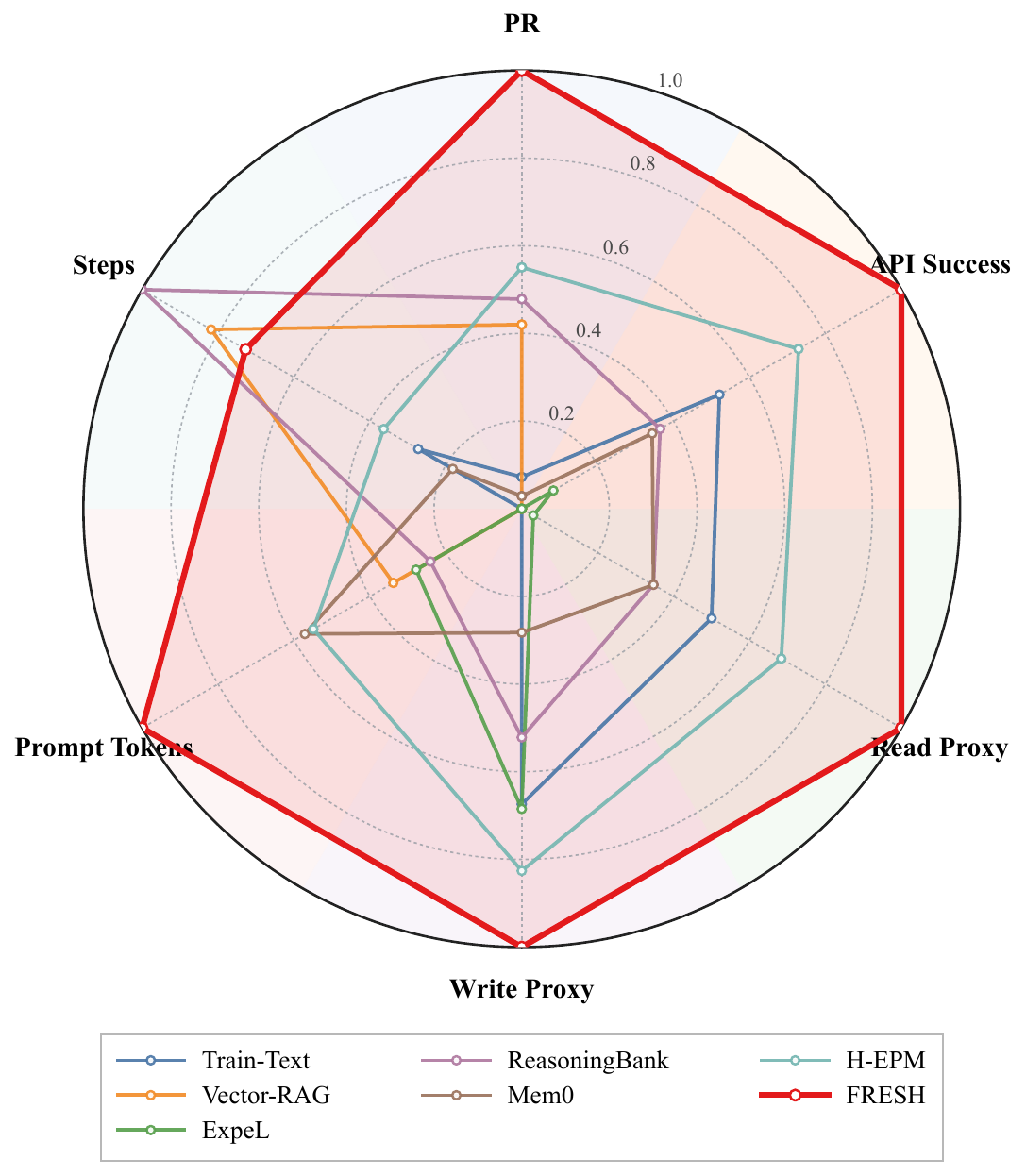}
        \caption{AppWorld Test-C}
        \label{fig:radar-appworld-testc}
    \end{subfigure}
    \vspace{-0.3cm}
    \caption{Normalized performance comparison across different methods. Larger normalized values indicate better performance on all axes.}
    \label{fig:radar_comparison}
\end{figure}

\begin{figure}[t]
    \centering
    \includegraphics[width=\linewidth]{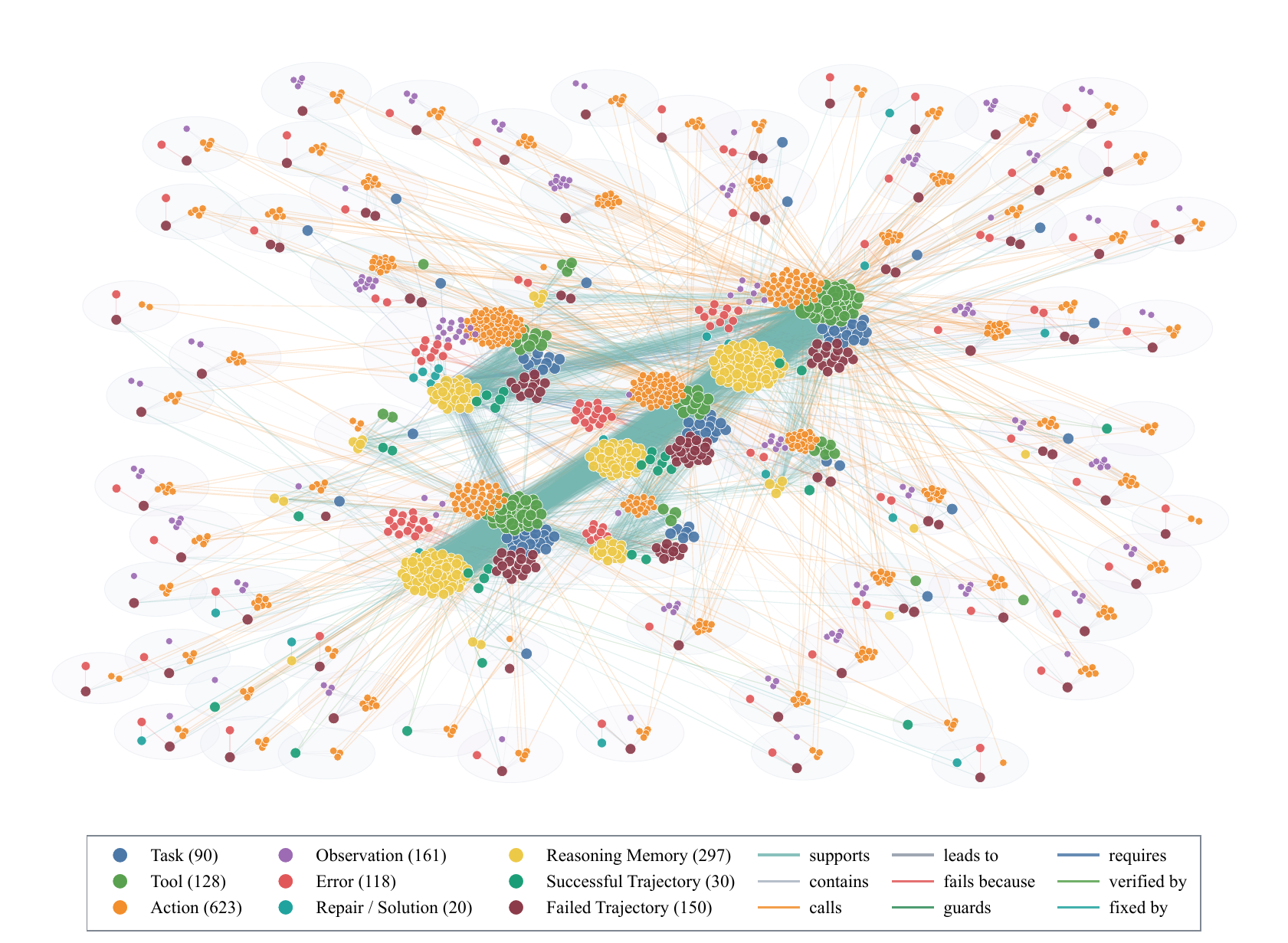}
    \vspace{-0.7cm}
    \caption{Visualization of a representative heterogeneous graph constructed from historical tool-use trajectories. 
    }
    \label{fig:experience_graph}
\end{figure}

\subsection{Reliability--Efficiency and Robustness Analysis}
We further analyze the reliability, resource efficiency, and challenge robustness of different memory methods. As shown in Figures~\ref{fig:performance--efficiency} and~\ref{fig:radar_comparison}, \method{} achieves a more favorable balance across these dimensions. In Figures~\ref{fig:performance--efficiency}, \method{} is positioned in the upper-right region and maintains strong challenge retention, indicating that its reliability gains do not come at the cost of substantially higher resource consumption. Figure~\ref{fig:radar_comparison} further shows that \method{} achieves the most comprehensive performance profile on both $\tau$-bench Airline and AppWorld Test-C, with clear advantages in task success and action-level reliability while retaining competitive prompt usage and execution steps. In contrast, the no-memory agent is resource-efficient but substantially less reliable, whereas existing memory methods generally improve only part of the evaluation dimensions. These results demonstrate that \method{} provides more reliable tool use through effective reuse of structured failure-aware experience rather than simply relying on longer contexts or additional interactions.

\subsection{Experience Graph Visualization}

Figure~\ref{fig:experience_graph} illustrates the structure of the experience memory learned by \method{}. The graph contains dense task-centered clusters that connect actions with their corresponding tools, observations, outcomes, and reasoning memories. Successful and failed trajectories are represented jointly, while error nodes are linked to their causes and possible repairs through explicit typed relations. The graph also contains cross-task connections that allow reusable tools, repair strategies, and execution conditions to be shared among related tasks. This structured organization enables \method{} to retrieve not only semantically similar experience, but also the dependencies and failure-recovery evidence needed for reliable execution.
\section{Conclusion}
We presented \method{}, a failure-aware heterogeneous graph memory framework for reliable tool-using agents. \method{} transforms successful and failed trajectories into structured experience that explicitly distinguishes verified actions, failure patterns, repairs, and execution preconditions. A graph-aware retriever provides compact operational guidance, while a lightweight policy and precondition gate constrains risky actions before execution. Experiments on $\tau$-bench and AppWorld demonstrate that \method{} consistently improves task completion and tool-use reliability across different small and medium-sized language models, while often reducing unnecessary execution steps. 



\bibliography{grace_refs}

\begin{thebibliography}{23}
\providecommand{\natexlab}[1]{#1}

\bibitem[{Brown et~al.(2020)Brown, Mann, Ryder, Subbiah, Kaplan, Dhariwal, Neelakantan, Shyam, Sastry, Askell, Agarwal, Herbert-Voss, Krueger, Henighan, Child, Ramesh, Ziegler, Wu, Winter, Hesse, Chen, Sigler, Litwin, Gray, Chess, Clark, Berner, McCandlish, Radford, Sutskever, and Amodei}]{brown2020language}
Brown, T.~B.; Mann, B.; Ryder, N.; Subbiah, M.; Kaplan, J.; Dhariwal, P.; Neelakantan, A.; Shyam, P.; Sastry, G.; Askell, A.; Agarwal, S.; Herbert-Voss, A.; Krueger, G.; Henighan, T.; Child, R.; Ramesh, A.; Ziegler, D.~M.; Wu, J.; Winter, C.; Hesse, C.; Chen, M.; Sigler, E.; Litwin, M.; Gray, S.; Chess, B.; Clark, J.; Berner, C.; McCandlish, S.; Radford, A.; Sutskever, I.; and Amodei, D. 2020.
\newblock Language Models are Few-Shot Learners.
\newblock In \emph{Advances in Neural Information Processing Systems}, volume~33, 1877--1901.

\bibitem[{Chhikara et~al.(2025)Chhikara, Khant, Aryan, Singh, and Yadav}]{mem02024}
Chhikara, P.; Khant, D.; Aryan, S.; Singh, T.; and Yadav, D. 2025.
\newblock Mem0: Building production-ready ai agents with scalable long-term memory.
\newblock \emph{arXiv preprint arXiv:2504.19413}.

\bibitem[{Feng et~al.(2026)Feng, Ye, Luo, Xu, Xu, Zhang, Hua, Xie, Yang, Liu, and You}]{feng2026expgraph}
Feng, T.; Ye, C.; Luo, T.; Xu, J.; Xu, X.; Zhang, H.; Hua, Z.; Xie, Y.; Yang, S.; Liu, G.; and You, J. 2026.
\newblock ExpGraph: Model-Agnostic Experience Learning with Graph-Structured Memory for LLM Agents.
\newblock arXiv:2605.30712.

\bibitem[{Karpas et~al.(2022)Karpas, Abend, Belinkov, Lenz, Lieber, Ratner, Shoham, Bata, Levine, Leyton-Brown et~al.}]{karpas2022mrkl}
Karpas, E.; Abend, O.; Belinkov, Y.; Lenz, B.; Lieber, O.; Ratner, N.; Shoham, Y.; Bata, H.; Levine, Y.; Leyton-Brown, K.; et~al. 2022.
\newblock MRKL Systems: A modular, neuro-symbolic architecture that combines large language models, external knowledge sources and discrete reasoning.
\newblock \emph{arXiv preprint arXiv:2205.00445}.

\bibitem[{Lewis et~al.(2020)Lewis, Perez, Piktus, Petroni, Karpukhin, Goyal, Kuttler, Lewis, Yih, Rockt{\"a}schel, Riedel, and Kiela}]{lewis2020retrieval}
Lewis, P.; Perez, E.; Piktus, A.; Petroni, F.; Karpukhin, V.; Goyal, N.; Kuttler, H.; Lewis, M.; Yih, W.-t.; Rockt{\"a}schel, T.; Riedel, S.; and Kiela, D. 2020.
\newblock Retrieval-Augmented Generation for Knowledge-Intensive {NLP} Tasks.
\newblock In \emph{Advances in Neural Information Processing Systems}, volume~33, 9459--9474.

\bibitem[{Li et~al.(2026)Li, Huang, Liu, Li, Fu, Song, Bian, Zhang, and Wang}]{li2026hepm}
Li, S.; Huang, Y.; Liu, Z.; Li, Z.; Fu, J.; Song, L.; Bian, J.; Zhang, J.; and Wang, R. 2026.
\newblock Experience-Evolving Multi-Turn Tool-Use Agent with Hybrid Episodic{\textendash}Procedural Memory.
\newblock In \emph{Forty-third International Conference on Machine Learning}.

\bibitem[{Ling et~al.(2026)Ling, Chen, Guan, Qu, Akbar, Gopinathan, and Cornejo}]{ling2026pace}
Ling, C.; Chen, P.; Guan, A.; Qu, J.; Akbar, S.~A.; Gopinathan, M.; and Cornejo, E. 2026.
\newblock PACE: Two-Timescale Self-Evolution for Small Language Model Agents.
\newblock arXiv:2605.23019.

\bibitem[{Liu et~al.(2024)Liu, Zhao, Iandola, Lai, Tian, Fedorov, Xiong, Chang, Shi, Krishnamoorthi et~al.}]{liu2024mobilellm}
Liu, Z.; Zhao, C.; Iandola, F.; Lai, C.; Tian, Y.; Fedorov, I.; Xiong, Y.; Chang, E.; Shi, Y.; Krishnamoorthi, R.; et~al. 2024.
\newblock Mobilellm: Optimizing sub-billion parameter language models for on-device use cases.
\newblock In \emph{Forty-first International Conference on Machine Learning}.

\bibitem[{Ouyang et~al.(2026)Ouyang, Yan, Hsu, Chen, Jiang, Wang, Han, Le, Daruki, Tang, Tirumalashetty, Lee, Rofouei, Lin, Han, Lee, and Pfister}]{ouyang2026reasoningbank}
Ouyang, S.; Yan, J.; Hsu, I.-H.; Chen, Y.; Jiang, K.; Wang, Z.; Han, R.; Le, L.~T.; Daruki, S.; Tang, X.; Tirumalashetty, V.; Lee, G.; Rofouei, M.; Lin, H.; Han, J.; Lee, C.-Y.; and Pfister, T. 2026.
\newblock ReasoningBank: Scaling Agent Self-Evolving with Reasoning Memory.
\newblock In \emph{International Conference on Learning Representations}.

\bibitem[{Packer et~al.(2024)Packer, Wooders, Lin, Fang, Patil, Stoica, and Gonzalez}]{packer2023memgpt}
Packer, C.; Wooders, S.; Lin, K.; Fang, V.; Patil, S.~G.; Stoica, I.; and Gonzalez, J.~E. 2024.
\newblock MemGPT: Towards LLMs as Operating Systems.
\newblock arXiv:2310.08560.

\bibitem[{Park et~al.(2023)Park, O'Brien, Cai, Morris, Liang, and Bernstein}]{park2023generative}
Park, J.~S.; O'Brien, J.~C.; Cai, C.~J.; Morris, M.~R.; Liang, P.; and Bernstein, M.~S. 2023.
\newblock Generative Agents: Interactive Simulacra of Human Behavior.
\newblock In \emph{Proceedings of the 36th Annual ACM Symposium on User Interface Software and Technology}.

\bibitem[{Patil et~al.(2025)Patil, Mao, Yan, Ji, Suresh, Stoica, and Gonzalez}]{patil2025bfcl}
Patil, S.~G.; Mao, H.; Yan, F.; Ji, C. C.-J.; Suresh, V.; Stoica, I.; and Gonzalez, J.~E. 2025.
\newblock The Berkeley Function Calling Leaderboard ({BFCL}): From Tool Use to Agentic Evaluation of Large Language Models.
\newblock In \emph{Forty-second International Conference on Machine Learning}.

\bibitem[{Patil et~al.(2024)Patil, Zhang, Wang, and Gonzalez}]{patil2024gorilla}
Patil, S.~G.; Zhang, T.; Wang, X.; and Gonzalez, J.~E. 2024.
\newblock Gorilla: Large language model connected with massive apis.
\newblock \emph{Advances in Neural Information Processing Systems}, 37: 126544--126565.

\bibitem[{Qin et~al.(2024)Qin, Liang, Ye, Zhu, Yan, Lu, Lin, Cong, Tang, Qian, Zhao, Hong, Tian, Xie, Zhou, Gerstein, Li, Liu, and Sun}]{qin2024toolllm}
Qin, Y.; Liang, S.; Ye, Y.; Zhu, K.; Yan, L.; Lu, Y.; Lin, Y.; Cong, X.; Tang, X.; Qian, B.; Zhao, S.; Hong, L.; Tian, R.; Xie, R.; Zhou, J.; Gerstein, M.; Li, D.; Liu, Z.; and Sun, M. 2024.
\newblock {ToolLLM}: Facilitating Large Language Models to Master 16000+ Real-World {API}s.
\newblock In \emph{International Conference on Learning Representations}.

\bibitem[{Schick et~al.(2023)Schick, Dwivedi-Yu, Dess{\`\i}, Raileanu, Lomeli, Hambro, Zettlemoyer, Cancedda, and Scialom}]{schick2023toolformer}
Schick, T.; Dwivedi-Yu, J.; Dess{\`\i}, R.; Raileanu, R.; Lomeli, M.; Hambro, E.; Zettlemoyer, L.; Cancedda, N.; and Scialom, T. 2023.
\newblock Toolformer: Language models can teach themselves to use tools.
\newblock \emph{Advances in neural information processing systems}, 36: 68539--68551.

\bibitem[{Shinn et~al.(2023)Shinn, Cassano, Gopinath, Narasimhan, and Yao}]{shinn2023reflexion}
Shinn, N.; Cassano, F.; Gopinath, A.; Narasimhan, K.; and Yao, S. 2023.
\newblock Reflexion: Language agents with verbal reinforcement learning.
\newblock \emph{Advances in neural information processing systems}, 36: 8634--8652.

\bibitem[{Team et~al.(2026)Team, Abd, Aggarwal, Algayres, Andreev, Bachem, Ballantyne, Brick, C{\u{a}}rbune, Casbon et~al.}]{gemmateam2026gemma4}
Team, G.; Abd, S.~E.; Aggarwal, V.; Algayres, R.; Andreev, A.; Bachem, O.; Ballantyne, I.; Brick, C.; C{\u{a}}rbune, V.; Casbon, M.; et~al. 2026.
\newblock Gemma 4 technical report.
\newblock \emph{arXiv preprint arXiv:2607.02770}.

\bibitem[{Trivedi et~al.(2024)Trivedi, Khot, Hartmann, Manku, Dong, Li, Gupta, Sabharwal, and Balasubramanian}]{trivedi2024appworld}
Trivedi, H.; Khot, T.; Hartmann, M.; Manku, R.; Dong, V.; Li, E.; Gupta, S.; Sabharwal, A.; and Balasubramanian, N. 2024.
\newblock {A}pp{W}orld: A Controllable World of Apps and People for Benchmarking Interactive Coding Agents.
\newblock In Ku, L.-W.; Martins, A.; and Srikumar, V., eds., \emph{Proceedings of the 62nd Annual Meeting of the Association for Computational Linguistics (Volume 1: Long Papers)}, 16022--16076. Bangkok, Thailand: Association for Computational Linguistics.

\bibitem[{Wang et~al.(2023)Wang, Xie, Jiang, Mandlekar, Xiao, Zhu, Fan, and Anandkumar}]{wang2023voyager}
Wang, G.; Xie, Y.; Jiang, Y.; Mandlekar, A.; Xiao, C.; Zhu, Y.; Fan, L.; and Anandkumar, A. 2023.
\newblock Voyager: An Open-Ended Embodied Agent with Large Language Models.
\newblock arXiv:2305.16291.

\bibitem[{Yang et~al.(2025)Yang, Li, Yang, Zhang, Hui, Zheng, Yu, Gao, Huang, Lv et~al.}]{yang2025qwen3}
Yang, A.; Li, A.; Yang, B.; Zhang, B.; Hui, B.; Zheng, B.; Yu, B.; Gao, C.; Huang, C.; Lv, C.; et~al. 2025.
\newblock {Qwen3} Technical Report.
\newblock \emph{arXiv preprint arXiv:2505.09388}.

\bibitem[{Yao et~al.(2024)Yao, Shinn, Razavi, and Narasimhan}]{yao2024taubench}
Yao, S.; Shinn, N.; Razavi, P.; and Narasimhan, K. 2024.
\newblock {$\tau$-bench}: A Benchmark for Tool-Agent-User Interaction in Real-World Domains.
\newblock arXiv:2406.12045.

\bibitem[{Yao et~al.(2023)Yao, Zhao, Yu, Du, Shafran, Narasimhan, and Cao}]{yao2023react}
Yao, S.; Zhao, J.; Yu, D.; Du, N.; Shafran, I.; Narasimhan, K.; and Cao, Y. 2023.
\newblock ReAct: Synergizing Reasoning and Acting in Language Models.
\newblock In \emph{International Conference on Learning Representations}.

\bibitem[{Zhao et~al.(2024)Zhao, Huang, Xu, Lin, Liu, and Huang}]{zhao2024expel}
Zhao, A.; Huang, D.; Xu, Q.; Lin, M.; Liu, Y.-J.; and Huang, G. 2024.
\newblock ExpeL: LLM agents are experiential learners.
\newblock In \emph{Proceedings of the Thirty-Eighth AAAI Conference on Artificial Intelligence and Thirty-Sixth Conference on Innovative Applications of Artificial Intelligence and Fourteenth Symposium on Educational Advances in Artificial Intelligence}. AAAI Press.
\newblock ISBN 978-1-57735-887-9.

\end{thebibliography}

\clearpage
\newpage

\section{Appendix}

\subsection{Illustrative Example of Failure-Aware Graph Induction}
\label{app:graph-induction-example}

We provide a simplified Airline example to illustrate how \method{} converts a failed interaction trajectory into structured and reusable graph memory.

\begin{freshpromptbox}{Example C: Failure-Aware Graph Induction from an Airline Trajectory}
\small

\promptsection{promptgreen}{\# User Request}

\emph{``Cancel my flight tomorrow if the reservation is eligible.''}

\promptsection{promptblue}{\# Raw Tool-Use Trajectory}

After normalizing tool names, arguments, and observations, the interaction is represented as

\begin{equation}
\begin{aligned}
x_i &= \text{cancel a reservation if eligible},\\
a_{i,1} &= \texttt{GetBooking}(\texttt{booking\_id}),\\
o_{i,1} &= \text{booking found and current status returned},\\
a_{i,2} &= \texttt{CancelBooking}(\texttt{booking\_id}),\\
o_{i,2} &= \text{cancellation rejected: eligibility unverified},\\
F_i &= \{\text{task failure, action-check failure,}\\
&\hspace{2.6em}\text{database state unchanged}\}.
\end{aligned}
\label{eq:graph-induction-example}
\end{equation}

\promptsection{promptyellow}{\# Action-Level Credit Assignment}

Although the complete trajectory fails, \method{} does not label every action as incorrect.

\begin{itemize}[leftmargin=1.3em,itemsep=2pt,topsep=2pt]
    \item \texttt{GetBooking} is retained as a \textbf{verified read}, because it successfully retrieves the current reservation state.
    \item \texttt{CancelBooking} is marked as a \textbf{failed write}, because it is executed before cancellation eligibility and user confirmation are verified.
\end{itemize}

This action-level distinction prevents useful steps in a failed trajectory from being discarded.

\promptsection{promptred}{\# Failure Knowledge Extracted}

From the premature cancellation, the experience miner extracts three complementary forms of knowledge:

\begin{itemize}[leftmargin=1.3em,itemsep=2pt,topsep=2pt]
    \item \textbf{Error:} cancellation was attempted without verified eligibility.
    \item \textbf{Precondition:} check the booking status, applicable fare rule, and required user confirmation.
    \item \textbf{Repair:} replace the premature write with a verified read--check--confirm--write sequence.
\end{itemize}

\noindent
The corrected execution pattern is

\begin{center}
\scriptsize
\texttt{GetBooking}\\[-1pt]
$\downarrow$\\[-1pt]
\texttt{CheckFareRule}\\[-1pt]
$\downarrow$\\[-1pt]
\texttt{ConfirmCancellation}\\[-1pt]
$\downarrow$\\[-1pt]
\texttt{CancelBooking}.
\end{center}

\promptsection{promptyellow}{\# Induced Heterogeneous Graph Fragment}

The corresponding graph fragment contains explicit semantic dependencies:

\begin{equation}
\begin{aligned}
\textsc{Task}
&\xrightarrow{\texttt{contains}}
\textsc{GetBooking},\\
\textsc{GetBooking}
&\xrightarrow{\texttt{verified\_by}}
\textsc{SuccessfulRead},\\
\textsc{CancelBooking}
&\xrightarrow{\texttt{fails\_because}}
\textsc{EligibilityNotVerified},\\
\textsc{CancelBooking}
&\xrightarrow{\texttt{requires}}
\textsc{CheckCancellationPolicy},\\
\textsc{CheckCancellationPolicy}
&\xrightarrow{\texttt{guards}}
\textsc{CancelBooking},\\
\textsc{EligibilityNotVerified}
&\xrightarrow{\texttt{fixed\_by}}
\textsc{ReadCheckConfirmWrite}.
\end{aligned}
\label{eq:induced-graph-fragment}
\end{equation}

Unlike a temporal trajectory chain, these typed relations explicitly describe which step succeeded, why the write failed, what condition was missing, and how the failure can be repaired.

\promptsection{promptblue}{\# Trajectory-to-Graph Conversion}

{\scriptsize
\setlength{\tabcolsep}{3pt}
\renewcommand{\arraystretch}{1.18}
\begin{tabularx}{\linewidth}{@{}>{\bfseries}p{0.24\linewidth}YY@{}}
\toprule
Trajectory signal & Induced graph memory & Subsequent use \\
\midrule
Conditional cancellation request
& Task node with cancellation intent and policy constraint
& Provides domain, intent, and constraint cues \\
\addlinespace
Successful booking lookup
& Verified action and observation linked by \texttt{verified\_by}
& Retrieved as a recommended read action \\
\addlinespace
Premature cancellation
& Failed action linked to an error by \texttt{fails\_because}
& Used as a negative sample and an \textsc{Avoid} pattern \\
\addlinespace
Missing eligibility check
& Precondition linked by \texttt{requires} and \texttt{guards}
& Retrieved as a \textsc{Check} condition \\
\addlinespace
Corrected execution order
& Repair linked by \texttt{fixed\_by} and \texttt{supports}
& Retrieved as a \textsc{Repair} strategy \\
\bottomrule
\end{tabularx}
}

\promptsection{promptgreen}{\# Serialized Guidance for a Future Task}

For a new reservation-cancellation request, the HG-Retriever can rank the verified read, missing precondition, and repair above the premature write. The selected graph evidence is serialized as:

\medskip
\noindent
\textbf{Do:} Retrieve the current booking state before planning the cancellation.

\smallskip
\noindent
\textbf{Avoid:} Do not cancel the reservation before eligibility is verified.

\smallskip
\noindent
\textbf{Check:} Validate the fare rule, booking status, and required user confirmation.

\smallskip
\noindent
\textbf{Repair:} If cancellation is not permitted, offer an allowed alternative or transfer the task.

\promptsection{promptblue}{\# Key Takeaway}

Failure-aware graph induction does not simply store the failed trajectory. It preserves the valid read action, suppresses the unsafe write, and converts the observed failure into reusable preconditions and recovery knowledge.

\end{freshpromptbox}

\begin{algorithm*}[t]
\caption{Overall Pipeline of \method{}}
\label{alg:fresh_overall}
\small
\begin{algorithmic}[1]
\Require Training/development trajectories $\mathcal{H}$, held-out tasks $\mathcal{D}_{\mathrm{test}}$, frozen executor $\pi_{\theta}$, tool set $\mathcal{T}$, domain policies $\mathcal{B}$, retrieval sizes $K_0$ and $K$
\Ensure Experience graph $\mathcal{G}$, trained retriever $\phi$, and task outputs

\State Initialize an empty heterogeneous graph $\mathcal{G}$
\For{each trajectory $\tau_i=(x_i,a_{i,1},o_{i,1},\ldots,a_{i,T_i},o_{i,T_i},F_i)\in\mathcal{H}$}
    \State Normalize tools, arguments, entities, and observations
    \State Perform action-level credit assignment using feedback $F_i$
    \State Add task, tool, action, observation, and trajectory nodes to $\mathcal{G}$
    \State Add verified strategies, failure causes, missing preconditions, and repairs with typed relations
\EndFor
\State Train the HGNN retriever $\phi\gets\Call{TrainHGNN}{\mathcal{G},\mathcal{H},K_0}$
\State Update graph utilities and insert newly observed failures or repairs using training/development feedback
\State Freeze $\mathcal{G}$ and $\phi$ before evaluation

\For{each held-out task $x\in\mathcal{D}_{\mathrm{test}}$}
    \State Initialize interaction history $h_0\gets[x]$
    \For{$t=1,\ldots,T_{\max}$}
        \State Construct query $\mathbf{q}_t\gets f_q(x,h_{t-1})$
        \State Recall and expand a candidate subgraph
        $\mathcal{G}^{c}_t\gets\Call{SparseRecall}{\mathcal{G},\mathbf{q}_t,K_0}$
        \State Rank candidate nodes with the HGNN and select top-$K$ memories
        \State Organize selected memories into \textsc{Do}, \textsc{Avoid}, \textsc{Check}, and \textsc{Repair}
        \State Serialize them as compact guidance $M_t$
        \State Generate a candidate action
        $\widetilde{a}_t\sim\pi_{\theta}(a\mid h_{t-1},M_t,\mathcal{T})$

        \If{$\widetilde{a}_t$ is state-changing or policy-sensitive}
            \State $d_t\gets\Call{PolicyGate}{\widetilde{a}_t,\mathcal{P}_t,\mathcal{O}_t,\mathcal{B}}$
            \State Convert $d_t\in\{\textsc{Allow},\textsc{Revise},\textsc{Block},\textsc{Transfer}\}$ into the executable action $a_t$
        \Else
            \State $a_t\gets\widetilde{a}_t$
        \EndIf

        \State Execute $a_t$, receive observation $o_t$, and update $h_t$
        \If{the task succeeds, terminates, or reaches the interaction budget}
            \State \textbf{break}
        \EndIf
    \EndFor
\EndFor
\State \Return $\mathcal{G}$, $\phi$, and task outputs
\end{algorithmic}
\end{algorithm*}

\begin{algorithm*}[t]
\caption{Training the Heterogeneous Graph Retriever}
\label{alg:hgnn_training}
\small
\begin{algorithmic}[1]
\Require Experience graph $\mathcal{G}$, historical trajectories $\mathcal{H}$, recall size $K_0$, neighborhood depth $H$, GNN layers $L$
\Ensure Trained retriever parameters $\phi$

\State Initialize ranking dataset $\mathcal{D}_{R}\gets\emptyset$

\For{each trajectory $\tau_i\in\mathcal{H}$}
    \State Construct structured query $\mathbf{q}_i=f_q(x_i,h_i)$
    \State Recall top-$K_0$ nodes and expand their typed $H$-hop neighbors to obtain $\mathcal{G}^{c}_i$
    \State Obtain positive nodes $\mathcal{V}^{+}_i$ from verified actions, successful strategies, repairs, and preconditions
    \State Obtain negative nodes $\mathcal{V}^{-}_i$ from failed actions, repeated calls, policy violations, and unsafe patterns
    \State Sample ranking pairs $(\mathbf{q}_i,\mathcal{G}^{c}_i,v^{+},v^{-})$ and add them to $\mathcal{D}_{R}$
\EndFor

\State Initialize HGNN parameters $\phi$
\For{each training epoch}
    \For{each mini-batch $\mathcal{B}\subset\mathcal{D}_{R}$}
        \State Construct node features from text, node type, historical statistics, and query matching
        \State Apply $L$ layers of relation-aware message passing over each candidate subgraph
        \State Compute query-conditioned node scores $s_{\phi}(\mathbf{q}_i,v)$
        \State Compute pairwise ranking loss
        \[
        \mathcal{L}_{\mathrm{rank}}
        =
        -\frac{1}{|\mathcal{B}|}
        \sum_{(\mathbf{q}_i,v^{+},v^{-})\in\mathcal{B}}
        \log\sigma\!\left(
        s_{\phi}(\mathbf{q}_i,v^{+})
        -
        s_{\phi}(\mathbf{q}_i,v^{-})
        \right)
        \]
        \State Compute
        $\mathcal{L}_{\mathrm{HG}}
        =
        \mathcal{L}_{\mathrm{rank}}
        +
        \lambda_{\mathrm{unsafe}}\mathcal{L}_{\mathrm{unsafe}}
        +
        \lambda_{\mathrm{reg}}\|\phi\|_2^2$
        \State Update $\phi$ by gradient descent
    \EndFor
\EndFor
\State Select the checkpoint on the development split and return $\phi$
\end{algorithmic}
\end{algorithm*}

\subsection{Datasets and Benchmarks}
\paragraph{$\tau$-bench.}
$\tau$-bench evaluates multi-turn interactions among a language-model agent, a simulated user, and stateful domain tools under explicit operational policies \citep{yao2024taubench}. The \textbf{Airline} domain contains tasks such as searching reservations, modifying flights, selecting seats, and processing cancellations, where the agent must verify booking states and fare policies before executing write operations. The \textbf{Retail} domain covers product search, order management, returns, exchanges, and refunds, requiring the agent to reason over user profiles, order states, product availability, and eligibility rules. The original benchmark contains 50 Airline tasks and 114 Retail tasks. Since it does not provide an official train--development--test partition, we construct fixed disjoint splits and reserve 20 Airline tasks and 40 Retail tasks for held-out evaluation. The remaining tasks are used only for collecting training and development trajectories, constructing the experience graph, and selecting hyperparameters. No held-out trajectory is used to update the graph or HG-Retriever.

Each test task is independently executed four times. We report $\mathrm{pass}^{k}$ for $k\in\{1,2,3,4\}$, which measures the probability that a task is successfully completed across $k$ sampled trials. While $\mathrm{pass}^{1}$ reflects average task success, larger $k$ provides a stricter measure of execution consistency. A run is considered successful only when the final database state and required actions satisfy the benchmark evaluator.

\paragraph{AppWorld.}
AppWorld evaluates long-horizon interactive coding agents in a controllable environment containing nine simulated daily applications, 457 APIs, and approximately 100 fictitious users \citep{trivedi2024appworld}. Its tasks require agents to inspect API documentation, execute multi-step code, maintain intermediate states, and coordinate operations across applications such as Amazon, Gmail, Spotify, Venmo, and Todoist. The benchmark contains 750 tasks derived from 250 task scenarios, with three task instances per scenario. The official partition consists of 105 training tasks, 60 development tasks, 168 normal test tasks (\textbf{Test-N}), and 417 challenging test tasks (\textbf{Test-C}). Test-C requires at least one API from an application unseen during training and is therefore more demanding in terms of API understanding and cross-application generalization.

We construct the AppWorld experience graph using only the training and development splits and freeze both the graph and HG-Retriever before evaluating on Test-N and Test-C. We report the task-level Pass Rate (PR), defined as the proportion of tasks satisfying the official state-based evaluation tests, together with the average number of agent execution steps. For additional diagnostic analysis, we report API-success, read-action, and write-action proxies, as well as prompt-token usage and inference cost.

\begin{table}[t]
\centering
\caption{Summary of the benchmark scales and evaluation splits used in our experiments.}
\label{tab:dataset_statistics}
\resizebox{\columnwidth}{!}{
\begin{tabular}{llrrl}
\toprule
\textbf{Benchmark} & \textbf{Domain/Split} & \textbf{Official Size} & \textbf{Test Size} & \textbf{Metrics} \\
\midrule
$\tau$-bench & Airline & 50 & 20 & $\mathrm{pass}^{1:4}$ \\
$\tau$-bench & Retail & 114 & 40 & $\mathrm{pass}^{1:4}$ \\
\midrule
AppWorld & Train & 105 & -- & Graph construction \\
AppWorld & Dev & 60 & -- & Model selection \\
AppWorld & Test-N & 168 & 168 & PR, Steps \\
AppWorld & Test-C & 417 & 417 & PR, Steps \\
\bottomrule
\end{tabular}
}
\end{table}

\subsection{Details on Baselines}

We compare \method{} with representative approaches covering prompting, retrieval, trajectory reuse, and experience learning. To ensure a fair comparison, all methods use the same training and development trajectories, held-out task splits, executor backbone, maximum interaction budget, and benchmark-specific system instructions. For retrieval-based methods, the number of retrieved memories and the memory prompt budget are matched to \method{} whenever applicable, while method-specific hyperparameters are selected on the development split.

\paragraph{No Memory.}
The no-memory agent directly solves each task using the frozen language-model executor, the current interaction history, domain policies, and available tool schemas. It does not access any historical trajectory or external experience and serves as the basic executor baseline.

\paragraph{Train-Text.}
Train-Text is a prompt-based baseline that converts historical training experience into a fixed textual prompt \citep{brown2020language}. The same prompt is provided to every test task without task-specific retrieval. This baseline evaluates whether simply exposing the executor to additional experience descriptions is sufficient to improve tool use.

\paragraph{Vector-RAG.}
Vector-RAG represents historical experience as independent textual memory entries and retrieves the most semantically similar entries using dense embedding similarity \citep{lewis2020retrieval}. The retrieved entries are appended to the executor context. Unlike \method{}, this baseline does not preserve typed dependencies among actions, observations, failures, preconditions, and repairs, and does not apply graph-based reranking.

\paragraph{Trajectory Replay.}
Trajectory Replay stores complete historical tool-use trajectories and retrieves task-relevant trajectories according to semantic similarity, following the general experiential-reuse setting of ExpeL \citep{zhao2024expel}. Retrieved trajectories are replayed as demonstrations until the memory context budget is reached. This baseline preserves detailed execution traces but may also introduce task-specific identifiers, irrelevant intermediate observations, and failed actions.

\paragraph{ReasoningBank.}
ReasoningBank distills successful and failed interactions into reusable reasoning memories and retrieves memories relevant to the current task \citep{ouyang2025reasoningbank}. It provides compact high-level strategies rather than replaying complete trajectories. However, its memories are represented as textual units and do not explicitly encode action dependencies or execution preconditions.

\paragraph{Mem0.}
Mem0 extracts concise long-term memories from historical interactions, consolidates related information, and performs semantic retrieval for subsequent tasks \citep{chhikara2025mem0}. We provide Mem0 with the same historical trajectory pool and insert its retrieved memories into the executor prompt under the same context-budget constraint.

\paragraph{H-EPM.}
H-EPM combines episodic memory with procedural tool-transition knowledge for multi-turn tool use \citep{li2026hepm}. It retrieves relevant historical episodes when task-specific context is needed and otherwise reuses high-confidence tool-transition patterns. Compared with \method{}, H-EPM primarily models successful episodic and procedural regularities, whereas \method{} additionally represents failure causes, repairs, write-action preconditions, and risky-action constraints.

\subsection{Details on SLM Backbones}

We evaluate all methods using three instruction-tuned open-source language models with different capacities. The model parameters remain frozen throughout graph construction, retriever training, and evaluation. Within each backbone, all compared methods use the same decoding configuration, context limit, tool schemas, maximum execution steps, and benchmark-specific system prompt. External memories are inserted into a dedicated memory block without modifying the original user request or tool definitions.

\paragraph{Llama-3.2-1B-Instruct.}
Llama-3.2-1B-Instruct\footnote{\url{https://huggingface.co/meta-llama/Llama-3.2-1B-Instruct}} is used as the lightweight executor. Its limited capacity makes it particularly suitable for evaluating whether external structured experience can compensate for weak long-horizon state tracking, planning, and error recovery.

\paragraph{Qwen3-8B.}
Qwen3-8B \citep{yang2025qwen3}\footnote{\url{https://huggingface.co/Qwen/Qwen3-8B}} serves as the intermediate-capacity executor. It provides stronger instruction following and reasoning ability than the 1B model while remaining practical for local or moderately resourced deployment.

\paragraph{Gemma-4-26B-A4B.}
Gemma-4-26B-A4B\footnote{\url{https://huggingface.co/google/gemma-4-26B-A4B}} is used as the relatively capable executor. It allows us to examine whether failure-aware graph memory remains beneficial when the underlying model already has stronger planning and tool-use capabilities. Across all three backbones, \method{} improves execution solely through external experience organization, graph retrieval, and action validation, without updating the language-model parameters.

\subsection{Example of an Instantiated FRESH Prompt}
\label{app:prompt-example}

The following example is instantiated from a real Retail evaluation episode (Task 5, Trial 0). We anonymize entity identifiers and omit repeated low-ranked candidates for presentation. No operational instruction is manually added or modified.

\begin{freshpromptbox}{Prompt A: Instantiated FRESH Prompt for a Retail Exchange Task}
\small

\promptsection{promptgreen}{\# Current Task}

I would like to exchange a water bottle and a desk lamp from a recent order.

\promptsection{promptgreen}{\# Task/State Signature}

\begin{itemize}[leftmargin=1.3em,itemsep=1pt,topsep=1pt]
    \item Domain hints: Retail.
    \item Intent cues: Exchange.
\end{itemize}

\promptsection{promptyellow}{\# Retrieved Success Strategy}

\textbf{Confidence: 0.71}

Start by gathering the missing account facts and then select the next tool according to live observations.

\medskip
\noindent\textbf{Previously successful tool pattern:}

\begin{center}
\small
\path{find_user_id_by_name_zip}\\
$\downarrow$\\
\path{get_user_details}\\
$\downarrow$\\
\path{get_order_details}\\
$\downarrow$\\
\path{get_product_details}\\
$\downarrow$\\
\path{exchange_delivered_order_items}
\end{center}

\promptsection{promptred}{\# Failure Pitfall}

\textbf{Confidence: 0.86}

Avoid open-ended tool-use loops. After each observation, identify one unresolved condition and perform exactly one targeted check or repair.

\promptsection{promptyellow}{\# Repair Rule}

\textbf{Confidence: 0.85}

After repeated turns, summarize the known facts, the latest observation, and the unresolved blocker. Then perform one verified action, ask one targeted question, or transfer the task when it cannot be safely completed.

\promptsection{promptred}{\# Policy/Precondition Gate}

Before executing a risky write action:

\begin{itemize}[leftmargin=1.3em,itemsep=1pt,topsep=1pt]
    \item Verify object ownership and the current order status.
    \item Verify policy eligibility and required user confirmation.
    \item Treat a database mismatch as evidence that the write sequence may be incomplete or inconsistent with the live state.
\end{itemize}

\promptsection{promptyellow}{\# Verified Tool/Action Pattern}

\noindent
\texttt{exchange\_delivered\_order\_items(}\\
\hspace*{1em}\texttt{item\_ids=[<ITEM\_ID>],}\\
\hspace*{1em}\texttt{new\_item\_ids=[<NEW\_ITEM\_ID>],}\\
\hspace*{1em}\texttt{order\_id=<ORDER\_ID>,}\\
\hspace*{1em}\texttt{payment\_method\_id=<PAYMENT\_ID>}\\
\texttt{)}

\medskip
\noindent
\textbf{Outcome:} success \qquad
\textbf{Confidence:} 0.67

\promptsection{promptred}{\# Failed Action Pattern to Avoid}

Do not reuse a historical exchange action unless its object identifiers, item variants, payment method, and eligibility conditions have been verified against the current environment state.

\promptsection{promptblue}{\# Runtime Requirements}

\begin{itemize}[leftmargin=1.3em,itemsep=1pt,topsep=1pt]
    \item Treat memory as guidance; current policies and live tool observations are authoritative.
    \item Allow a risky write only after live facts and explicit confirmation have been verified.
    \item Revise the plan to a read action or targeted question when required information is missing.
    \item Never allow a retrieved success path to bypass its preconditions.
\end{itemize}

\end{freshpromptbox}

\begin{table*}[t]
    \centering
    \caption{Performance and efficiency comparison on the $\tau$-bench Airline domain.
    Higher values are better for task performance and action accuracy, while lower values
    are better for prompt tokens and average cost. The best results are in \textbf{bold},
    and the second-best results are \underline{underlined}.}
    \label{tab:airline_cost_analysis}
    \resizebox{\textwidth}{!}{
    \begin{tabular}{lcccccccc}
        \toprule
        \multirow{2}{*}{Method}
        & \multicolumn{2}{c}{Task Performance}
        & \multicolumn{3}{c}{Action Accuracy (\%)}
        & \multicolumn{2}{c}{Efficiency} \\
        \cmidrule(lr){2-3}
        \cmidrule(lr){4-6}
        \cmidrule(lr){7-8}
        & $\mathrm{Pass}^{1}\uparrow$
        & $\mathrm{Pass}^{4}\uparrow$
        & Tool/Action $\uparrow$
        & Read $\uparrow$
        & Write $\uparrow$
        & Prompt Tokens $\downarrow$
        & Avg. Cost $\downarrow$ \\
        \midrule

        No-memory
        & 0.4875
        & \underline{0.3500}
        & 42.70
        & 64.60
        & 11.80
        & \textbf{59.5k}
        & \$0.005383 \\

        Train-Text
        & \underline{0.5250}
        & 0.2500
        & 38.80
        & 61.50
        & 3.60
        & 186.5k
        & \$0.016970 \\

        Vector-RAG
        & 0.4500
        & 0.2000
        & 47.80
        & 75.50
        & 7.70
        & 95.4k
        & \$0.008823 \\

        ExpeL
        & 0.4750
        & 0.2500
        & \underline{58.10}
        & \underline{82.30}
        & 9.70
        & 81.2k
        & \$0.007638 \\

        ReasoningBank
        & 0.3500
        & 0.2000
        & 46.30
        & 74.70
        & 7.00
        & 72.2k
        & \$0.006776 \\

        Mem0
        & 0.4375
        & \underline{0.3500}
        & 48.60
        & 73.00
        & \underline{8.60}
        & \underline{77.9k}
        & \$0.005660 \\

        H-EPM
        & 0.4625
        & 0.250
        & 50.00
        & 75.30
        & \underline{15.90}
        & \underline{63.3k}
        & \$0.008119 \\

        \midrule
        \rowcolor{gray!12}
        \textbf{FRESH (Ours)}
        & \textbf{0.5375}
        & \textbf{0.4500}
        & \textbf{76.20}
        & \textbf{90.40}
        & \textbf{54.40}
        & 70.0k
        & \textbf{\$0.005956} \\

        \bottomrule
    \end{tabular}
    }
\end{table*}
\begin{table*}[t]
\centering

\caption{Performance and efficiency comparison on the
$\tau$-bench Retail domain. Higher values are better for task
performance and action accuracy, while lower values are better
for prompt tokens and average cost. The best results are in
\textbf{bold}, and the second-best results are \underline{underlined}.}
\label{tab:retail-main}
\resizebox{\textwidth}{!}{
\begin{tabular}{lccccccc}
\toprule
& \multicolumn{2}{c}{Task Performance}
& \multicolumn{3}{c}{Action Accuracy (\%)}
& \multicolumn{2}{c}{Efficiency} \\
\cmidrule(lr){2-3}\cmidrule(lr){4-6}\cmidrule(lr){7-8}
Method
& Pass$^1$ $\uparrow$
& Pass$^4$ $\uparrow$
& Tool/Action $\uparrow$
& Read $\uparrow$
& Write $\uparrow$
& Prompt Tokens $\downarrow$
& Avg.\ Cost $\downarrow$ \\
\midrule
No-memory
& 0.6687 & 0.4500 & \underline{90.05}
& \textbf{94.51} & 78.31 & \textbf{76.0k} & \textbf{\$0.006852} \\
Train-Text
& \underline{0.6750} & 0.4500 & 88.56
& 92.18 & 79.25 & 258.4k & \$0.019231 \\
Vector-RAG
& 0.6687 & \underline{0.5000} & 88.66
& 91.79 & \underline{80.41} & 111.6k & \$0.010306 \\
ExpeL
& 0.6625 & 0.500 & 89.18
& 93.02 & 79.19 & 98.9k & \$0.008868 \\
ReasoningBank
& 0.6438 & 0.4750 & 88.44
& \underline{93.85} & 72.97 & \underline{87.3k}
& \underline{\$0.008007} \\
Mem0
& 0.6562 & 0.4750 & 88.55
& 93.65 & 75.00 & 109.3k & \$0.010109 \\
H-EPM
& 0.6188 & 0.4500 & 88.93
& 93.19 & 76.92 & 102.4k & \$0.009929 \\
\midrule
\rowcolor{gray!10}
\textbf{FRESH (Ours)}
& \textbf{0.7000} & \textbf{0.5250} & \textbf{90.15}
& 93.07 & \textbf{81.94} & 107.3k & \$0.009592 \\
\bottomrule
\end{tabular}
}
\end{table*}

\begin{table*}[t]
    \centering
    \caption{Performance and efficiency comparison on the AppWorld
    Test-N split. Higher values are better for PR and action proxies,
    while lower values are better for prompt tokens and average cost.
    The best results are in \textbf{bold}, and the second-best results
    are \underline{underlined}.}
    \label{tab:appworld_test_n}
    \resizebox{\textwidth}{!}{
    \begin{tabular}{lcccccc}
        \toprule
        Method
        & \makecell{PR $\uparrow$}
        & \makecell{API Success Proxy $\uparrow$}
        & \makecell{Read Proxy $\uparrow$}
        & \makecell{Write Proxy $\uparrow$}
        & \makecell{Prompt Tokens $\downarrow$}
        & \makecell{Avg. Cost $\downarrow$} \\
        \midrule

        No-Memory
        & 0.131
        & 40.4
        & 40.7
        & 35.2
        & \textbf{94.0K}
        & \textbf{\$0.00938} \\

        Train-Text
        & \underline{0.250}
        & 55.2
        & 54.9
        & \underline{59.7}
        & 102.3K
        & \$0.00967 \\

        Vector-RAG
        & 0.202
        & 48.3
        & 48.7
        & 44.1
        & 102.2K
        & \$0.00983 \\

        ExpeL
        
        & 0.220
        & 44.9
        & 44.5
        & 49.8
        & 106.1K
        & \$0.00978 \\

        ReasoningBank
        & 0.244
        & 51.5
        & 51.4
        & 52.7
        & 101.6K
        & \$0.00981 \\

        Mem0
        & 0.173
        & 56.3
        & 56.0
        & \textbf{60.7}
        & 99.3K
        & \underline{\$0.00939} \\

        H-EPM
        & 0.250
        & \underline{58.1}
        & \underline{58.4}
        & \textbf{54.0}
        & 101.1K
        & \underline{\$0.00939} \\

        \midrule
        \rowcolor{gray!12}
        \textbf{FRESH}
        & \textbf{0.268}
        & \textbf{58.5}
        & \textbf{58.6}
        & 57.1
        & \underline{95.2K}
        & \$0.00981 \\

        \bottomrule
    \end{tabular}
    }
\end{table*}

\begin{table*}[t]
    \centering
    \caption{Performance and efficiency comparison on the AppWorld
    Test-C split. Higher values are better for PR and action proxies,
    while lower values are better for prompt tokens and average cost.
    The best results are in \textbf{bold}, and the second-best results
    are \underline{underlined}.}
    \label{tab:appworld_test_c}
    \resizebox{\textwidth}{!}{
    \begin{tabular}{lcccccc}
        \toprule
        Method
        & \makecell{PR $\uparrow$}
        & \makecell{API Success Proxy $\uparrow$}
        & \makecell{Read Proxy $\uparrow$}
        & \makecell{Write Proxy $\uparrow$}
        & \makecell{Prompt Tokens $\downarrow$}
        & \makecell{Avg. Cost $\downarrow$} \\
        \midrule

        No-Memory
        & 0.043
        & 28.8
        & 29.0
        & 25.9
        & \textbf{97.9K}
        & \textbf{\$0.00992} \\

        Train-Text
        & 0.082
        & \underline{38.7}
        & \underline{39.1}
        & 34.7
        & 116.5K
        & \$0.01128 \\

        Vector-RAG
        & 0.106
        & 33.7
        & 34.2
        & 28.5
        & 112.0K
        & \$0.01057 \\

        ExpeL
        & 0.077
        & 34.5
        & 34.5
        & \underline{34.8}
        & 112.8K
        & \$0.01055 \\

        ReasoningBank
        & \underline{0.110}
        & 37.2
        & 37.6
        & 33.3
        & 113.3K
        & \$0.01079 \\

        Mem0
        & 0.079
        & 37.0
        & 37.6
        & 31.1
        & 108.9K
        & \underline{\$0.01028} \\

        H-EPM
        & 0.115
        & \underline{40.7}
        & \underline{40.9}
        & \textbf{36.1}
        & 109.2K
        & \underline{\$0.01072} \\

        \midrule
        \rowcolor{gray!12}
        \textbf{GRACE}
        
         & \textbf{0.146}
        & \textbf{43.3}
        & \textbf{44.0}
        & \textbf{37.7}
        & \underline{103.2K}
        & \$0.01037 \\

        \bottomrule
    \end{tabular}
    }
\end{table*}

\begin{figure*}[t]
    \centering

    \begin{subfigure}[t]{0.32\textwidth}
        \centering
        \includegraphics[width=\linewidth]{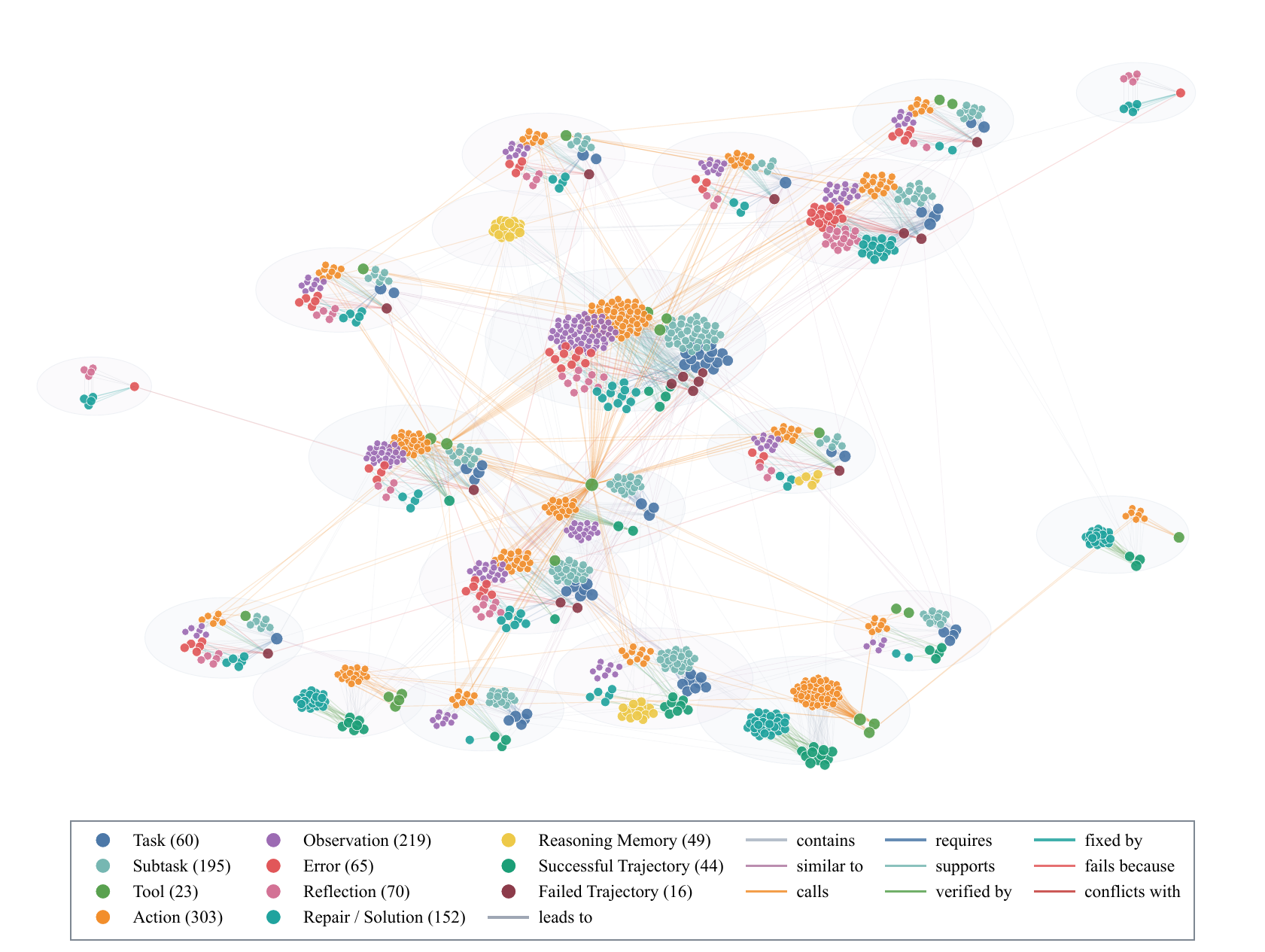}
        \caption{$\tau$-bench Airline}
        \label{fig:radar-tau-airline}
    \end{subfigure}
    \hfill
    \begin{subfigure}[t]{0.32\textwidth}
        \centering
        \includegraphics[width=\linewidth]{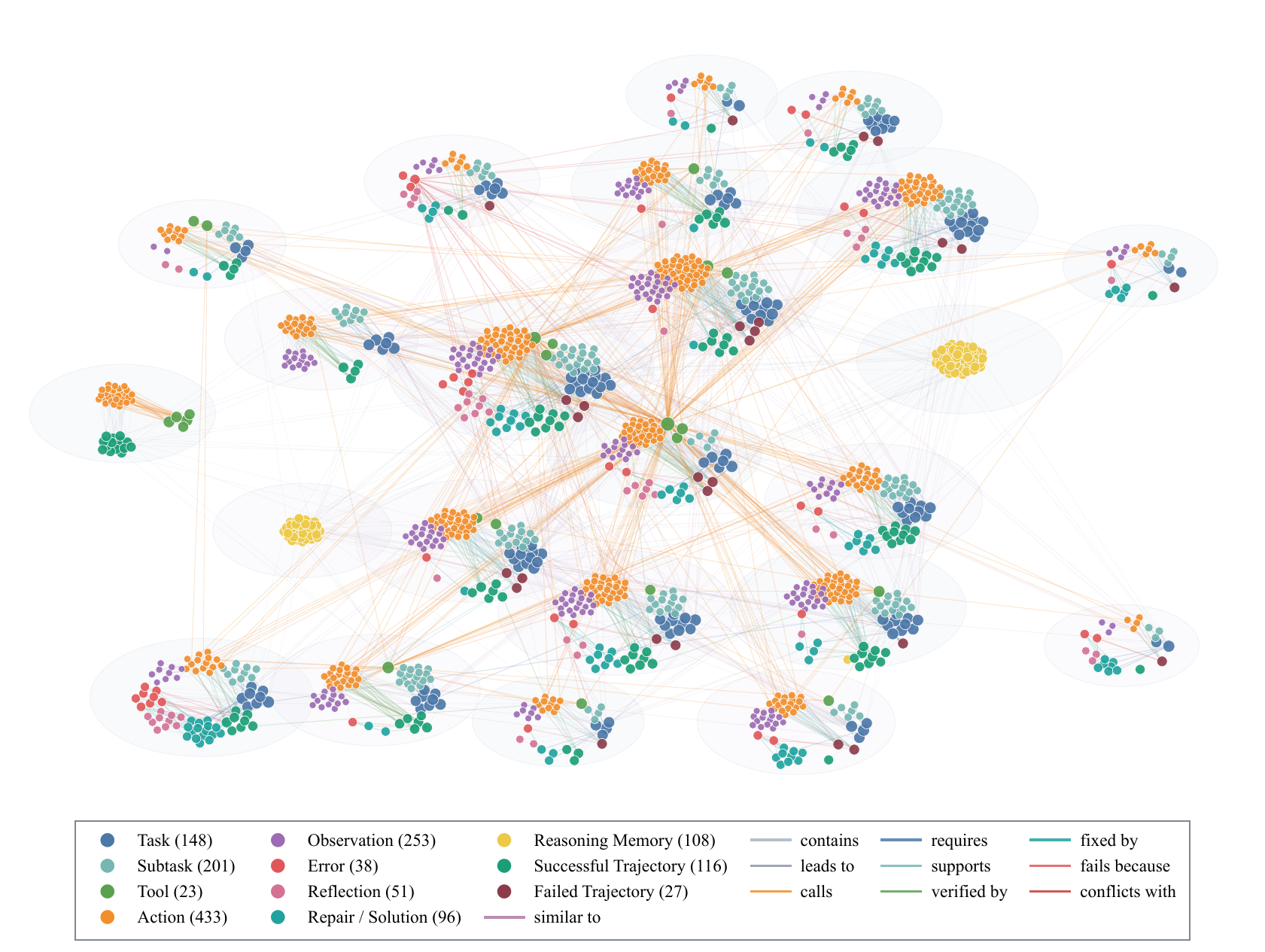}
        \caption{AppWorld Test-N}
        \label{fig:radar-appworld-testn}
    \end{subfigure}
    \hfill
    \begin{subfigure}[t]{0.32\textwidth}
        \centering
        \includegraphics[width=\linewidth]{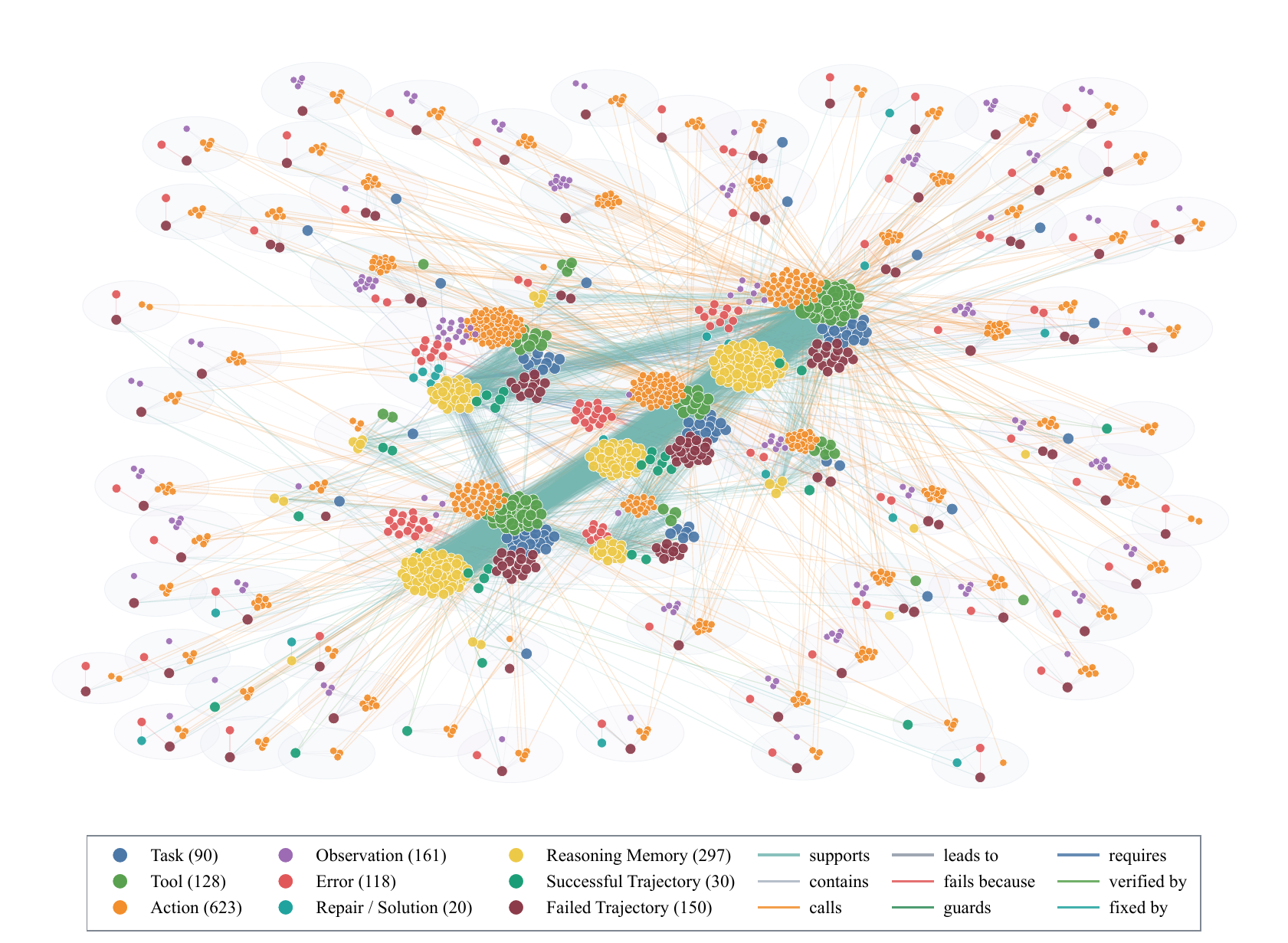}
        \caption{AppWorld Test-C}
        \label{fig:radar-appworld-testc}
    \end{subfigure}
    \vspace{-0.3cm}
    \caption{Normalized performance comparison across different methods. Larger normalized values indicate better performance on all axes.}
    \label{fig:graph}
\end{figure*}

\begin{figure*}[t]
    \centering
    \begin{subfigure}[t]{0.32\textwidth}
        \centering
        \includegraphics[width=\linewidth]{data/Figures/radar/radar_tau_bench_airline.pdf}
        \caption{$\tau$-bench Airline}
        \label{fig:radar-tau-airline}
    \end{subfigure}
    \hfill
    \begin{subfigure}[t]{0.32\textwidth}
        \centering
        \includegraphics[width=\linewidth]{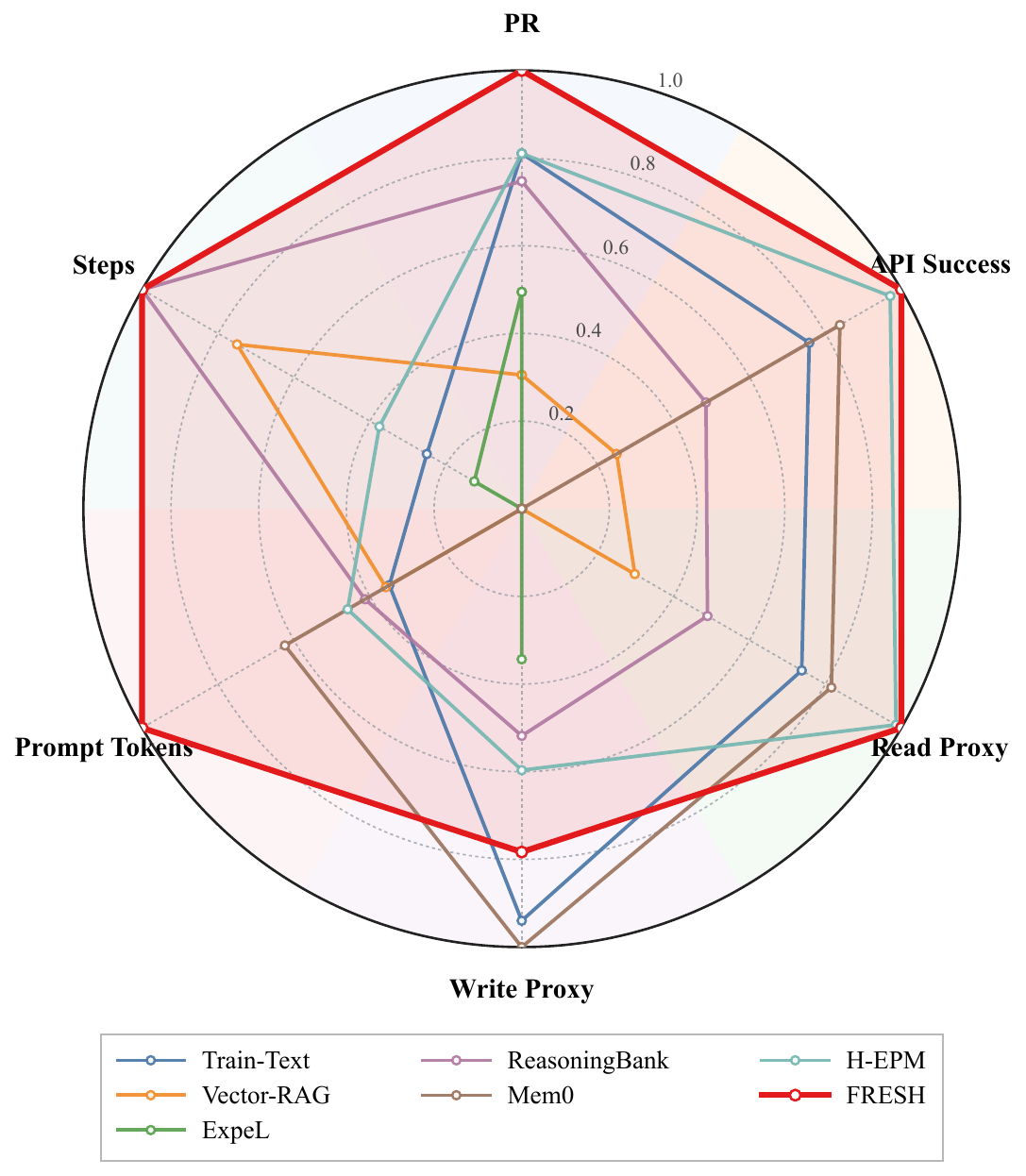}
        \caption{AppWorld Test-N}
        \label{fig:radar-appworld-testn}
    \end{subfigure}
    \hfill
    \begin{subfigure}[t]{0.32\textwidth}
        \centering
        \includegraphics[width=\linewidth]{data/Figures/radar/radar_appworld_testc.pdf}
        \caption{AppWorld Test-C}
        \label{fig:radar-appworld-testc}
    \end{subfigure}
    \vspace{-0.3cm}
    \caption{Normalized performance comparison across different methods. Larger normalized values indicate better performance on all axes.}
    \label{fig:radar_comparison2}
\end{figure*}

\subsection{Instantiated AppWorld Prompt}
\label{app:appworld-prompt}

The following prompt is reconstructed from a successful AppWorld Test-N episode (\texttt{fd1f8fa\_3}) using the same task specification, training graph, R-GCN checkpoint, and prompt-construction code. We anonymize credentials and omit repetitive API documentation.

\begin{freshpromptbox}{Prompt B: Instantiated FRESH Prompt for AppWorld}
\small

\promptsection{promptgreen}{\# System Role}

You are an AppWorld code agent. At every turn, output exactly one Python code block and no additional prose.

\promptsection{promptgreen}{\# Execution Environment}

The code runs in a persistent Python shell with \texttt{apis} and \texttt{requester} available. Inspect the exact API documentation before guessing parameters. After a schema or authentication error, inspect the corresponding API documentation and retry with corrected arguments.

\promptsection{promptblue}{\# General Execution Requirements}

\begin{itemize}[leftmargin=1.3em,itemsep=1pt,topsep=1pt]
    \item Use each turn for a compact batch of useful API calls.
    \item Avoid repeating failed calls or broad documentation searches.
    \item Call \texttt{apis.supervisor.complete\_task(...)} only after the requested state has been verified.
    \item Do not access the operating-system filesystem, external network, or hidden ground truth.
\end{itemize}

\promptsection{promptyellow}{\# FRESH Experience Memory}

The following hints are distilled only from training and development trajectories. They should be treated as execution guidance rather than task-specific ground truth.

\promptsection{promptblue}{\# Memory Usage Constraints}

\begin{itemize}[leftmargin=1.3em,itemsep=1pt,topsep=1pt]
    \item Do not copy historical IDs, emails, passwords, tokens, amounts, or dates.
    \item The current instruction, API documentation, and live observations override retrieved memory.
    \item Use memory to determine the next verification step rather than replaying a complete historical trajectory.
\end{itemize}

\promptsection{promptgreen}{\# Universal AppWorld Discipline}

\begin{itemize}[leftmargin=1.3em,itemsep=1pt,topsep=1pt]
    \item Inspect exact API documentation after schema or authentication errors.
    \item Parse supervisor passwords as a list of \texttt{\{account\_name, password\}} records.
    \item Verify the final application state before calling \texttt{complete\_task}.
    \item Avoid repeated broad searches or unnecessary documentation calls.
\end{itemize}

\promptsection{promptyellow}{\# Relevant Observed APIs}

\begin{itemize}[leftmargin=1.3em,itemsep=1pt,topsep=1pt]
    \item \texttt{api\_docs.search\_api\_docs} \hfill [read, confidence=0.61]
    \item \texttt{api\_docs.show\_api\_doc} \hfill [read, confidence=0.61]
    \item \texttt{supervisor.show\_account\_passwords} \hfill [read, confidence=0.68]
    \item \texttt{supervisor.show\_profile} \hfill [read, confidence=0.49]
    \item \texttt{spotify.login} \hfill [read, confidence=0.65]
    \item \texttt{spotify.show\_song\_library} \hfill [read, confidence=0.55]
    \item \texttt{spotify.show\_downloaded\_songs} \hfill [read, confidence=0.51]
    \item \texttt{supervisor.complete\_task} \hfill [write, confidence=0.70]
\end{itemize}

\promptsection{promptyellow}{\# Verified Success Patterns}

For Spotify queue-management tasks, successful trajectories generally authenticate the user, inspect the current queue and liked-song state, perform the required mutations, verify the resulting queue and player state, and finally call \texttt{supervisor.complete\_task}.

\medskip
\noindent\textbf{Previously successful tool sequence:}

\begin{center}
\scriptsize
\texttt{supervisor.show\_account\_passwords}\\
$\downarrow$\\
\texttt{api\_docs.show\_api\_doc}\\
$\downarrow$\\
\texttt{supervisor.show\_profile}\\
$\downarrow$\\
\texttt{spotify.login}\\
$\downarrow$\\
\texttt{spotify.show\_song\_queue}\\
$\downarrow$\\
\texttt{Spotify mutation APIs}\\
$\downarrow$\\
\texttt{supervisor.complete\_task}
\end{center}

\promptsection{promptred}{\# Failure Patterns to Avoid}

\begin{itemize}[leftmargin=1.3em,itemsep=1pt,topsep=1pt]
    \item Do not repeat \texttt{show\_account\_passwords} after credentials have already been obtained.
    \item Do not repeatedly inspect downloaded songs when the task concerns liked songs.
    \item Do not perform broad library cleanup before identifying the songs currently present in the queue.
    \item Do not call \texttt{complete\_task} before verifying the queue and player state.
\end{itemize}

\promptsection{promptred}{\# Preconditions Before Task Completion}

Before calling \texttt{complete\_task}, verify that:

\begin{itemize}[leftmargin=1.3em,itemsep=1pt,topsep=1pt]
    \item all liked songs have been removed from the current queue;
    \item unrelated queue entries remain unchanged; and
    \item the Spotify player has been started.
\end{itemize}

\promptsection{promptblue}{\# Recommended Runtime Pattern}

Prefer compact API batches that authenticate the account, read the current state, perform the required mutations, verify the updated state, and then complete the task. Use FRESH memory only as a hint; the current AppWorld task and API documentation remain authoritative.

\promptsection{promptgreen}{\# User Task}

\textbf{Task ID:} \texttt{fd1f8fa\_3}

\textbf{Current datetime:} 2023-05-18 12:00:00

\medskip
\noindent\textbf{Instruction:}

Remove all the songs that I have liked from my Spotify queue, and then start the player.

\promptsection{promptyellow}{\# Allowed Applications}

\texttt{api\_docs}, \texttt{supervisor}, \texttt{spotify}, \texttt{phone}, \texttt{amazon}, \texttt{venmo}, \texttt{gmail}, \texttt{splitwise}, \texttt{simple\_note}, \texttt{todoist}, and \texttt{file\_system}.

\promptsection{promptyellow}{\# Available API Documentation}

\noindent\textbf{Supervisor APIs}

\begin{itemize}[leftmargin=1.3em,itemsep=1pt,topsep=1pt]
    \item \texttt{show\_profile}: Show the main user's profile.
    \item \texttt{show\_account\_passwords}: Return account-password records.
    \item \texttt{complete\_task}: Mark the task complete after verification.
\end{itemize}

\noindent\textbf{Spotify APIs}

\begin{itemize}[leftmargin=1.3em,itemsep=1pt,topsep=1pt]
    \item \texttt{login}: Log in to the Spotify account.
    \item \texttt{show\_song\_queue}: Return songs in the current queue.
    \item \texttt{show\_liked\_songs}: Return songs liked by the user.
    \item \texttt{remove\_song\_from\_queue}: Remove a song from the queue.
    \item \texttt{start\_player}: Start the Spotify player.
    \item \texttt{show\_player}: Return the current player state.
\end{itemize}

\promptsection{promptblue}{\# Reusable Authentication Pattern}

\begin{lstlisting}[style=freshcode]
profile = apis.supervisor.show_profile()
passwords = apis.supervisor.show_account_passwords()
password = next(
    p["password"] for p in passwords
    if p["account_name"] == "spotify"
)
\end{lstlisting}

\promptsection{promptblue}{\# Output Requirement}

Write Python code for the next AppWorld interaction.

\end{freshpromptbox}

\end{document}